\documentclass[sigconf, nonacm=true, natbib=false, review=false]{acmart}

\author{Jake Grigsby}
\affiliation{
\institution{University of Virginia\footnotemark}
}
\thanks{\footnotemark Work done in part while an intern at IBM Research. Now at UT Austin, email: \texttt{grigsby@cs.utexas.edu}}
\email{jcg6dn@virginia.edu}

\author{Zhe Wang}
\affiliation{
\institution{University of Virginia}
\country{}
}
\email{zw6sg@virginia.edu}

\author{Nam Nguyen}
\affiliation{
\institution{IBM Research}
\country{}
}
\email{nnguyen@us.ibm.com}

\author{Yanjun Qi}
\affiliation{
\institution{University of Virginia}
\country{}
}
\email{yanjun@virginia.edu}

\usepackage{soul}
\usepackage{amsmath}
\usepackage{mathtools}
\usepackage{biblatex}
\usepackage{amsthm}
\usepackage{microtype}
\usepackage{graphicx}
\usepackage{subfigure}
\usepackage{booktabs} 

\usepackage{graphicx,subfigure,epsfig,fancybox} 
\usepackage{float}
\usepackage{multirow}
\usepackage{bbm}
\usepackage{hyperref}
\hypersetup{
    colorlinks=true,
    linkcolor=blue,
    citecolor=cyan,
    filecolor=magenta,      
    urlcolor=cyan,
    }

\usepackage{adjustbox}
\usepackage{array}
\usepackage{booktabs}

\newcommand{\method}[0]{\texttt{Spacetimeformer}}

\usepackage{enumitem}
\setlist[itemize]{leftmargin=*}

\usepackage{enumitem}

\newcommand{\cref}[1]{Condition~(\ref{#1})}

\usepackage{pifont}
\newcommand{\xmark}{\ding{55}}%

\title{Long-Range Transformers for Dynamic Spatiotemporal Forecasting}

\begin{document}

\begin{abstract}

Multivariate time series forecasting focuses on predicting future values based on historical context. State-of-the-art sequence-to-sequence models rely on neural attention between timesteps, which allows for \textit{temporal} learning but fails to consider distinct \textit{spatial} relationships between variables. In contrast, methods based on graph neural networks explicitly model variable relationships. However, these methods often rely on predefined graphs that cannot change over time and perform separate spatial and temporal updates without establishing direct connections between each variable at every timestep. Our work addresses these problems by translating multivariate forecasting into a ``spatiotemporal sequence" formulation where each Transformer input token represents the value of a single variable at a given time. Long-Range Transformers can then learn interactions between space, time, and value information jointly along this extended sequence. Our method, which we call \method, achieves competitive results on benchmarks from traffic forecasting to electricity demand and weather prediction while learning spatiotemporal relationships purely from data. 

\end{abstract}

\maketitle

\section{Introduction}

\begin{figure}[h!]
    \centering
    \includegraphics[width=\columnwidth]{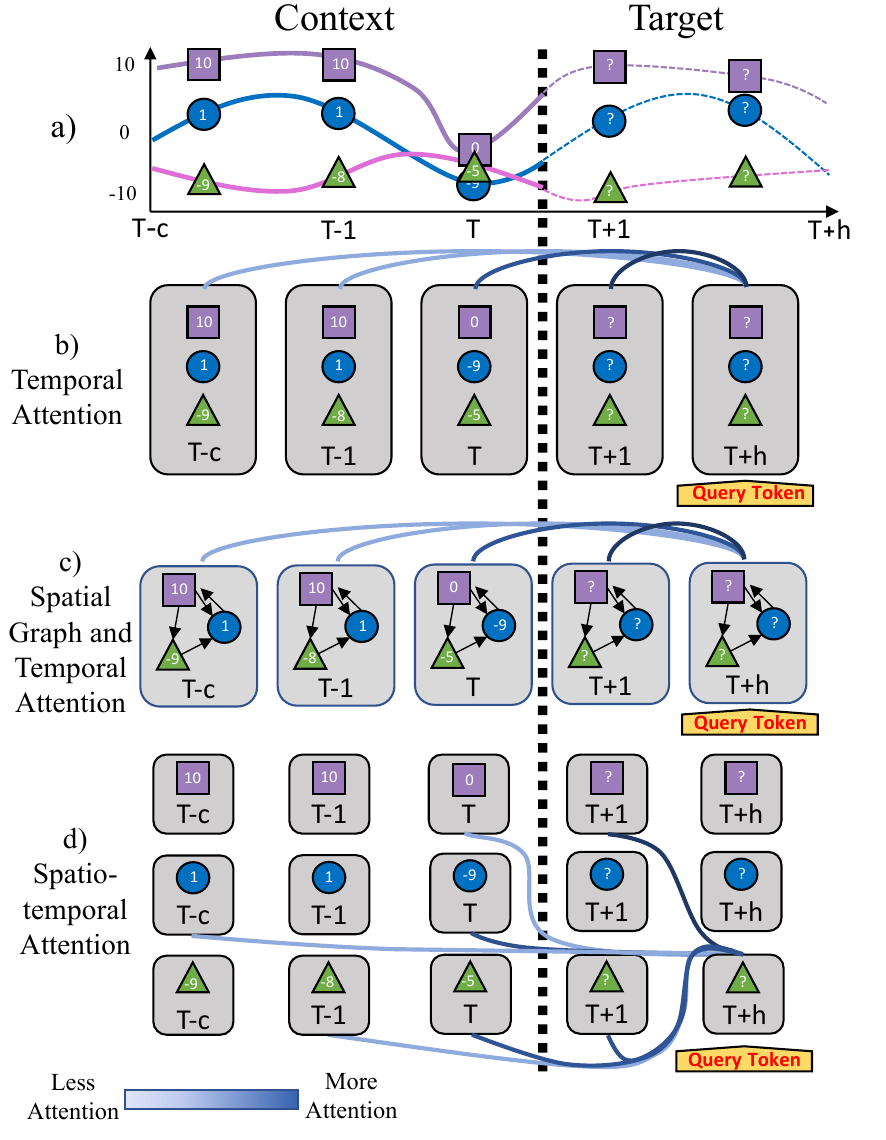}
    \caption{\textbf{Attention in multivariate forecasting.} (a) A three-variable sequence with three context points and two target points to predict. (b) Temporal attention in which each token contains all three variables. Darker blue lines between grey tokens represent increasing attention. (c) Temporal attention with spatial interactions modeled within each token's timestep by known spatial graphs (shown in black). (d) Spatiotemporal attention in which each variables at each timestep is a separate token. In practice this graph is densely connected but edges have been cut for readability. All figures in this paper are best viewed in color.}
    \label{fig:st-graph}
\end{figure}

Multivariate forecasting models attempt to predict future outcomes based on historical context; jointly modeling a set of variables allows us to interpret dependency relationships that provide early warning signs of changes in future behavior. A simple example is shown in Figure \ref{fig:st-graph}a. Time Series Forecasting (TSF) models typically deal with a small number of variables with long-term \textit{temporal} dependencies that require historical recall and distant forecasting. This is commonly handled by encoder-decoder sequence-to-sequence (seq2seq) architectures based on recurrent networks or one-dimensional convolutions. Current state-of-the-art TSF models substitute classic seq2seq architectures for neural-attention-based mechanisms. However, these models represent the value of multiple variables per timestep as a single input token. This lets them learn ``temporal attention" amongst timesteps but can ignore the distinct \textit{spatial} relationships that exist between variables. A temporal attention network is shown in Figure \ref{fig:st-graph}b.

In contrast, spatial-temporal methods aim to capture the relationships between multiple variables. These models typically involve alternating applications of temporal sequence processing and spatial message passing based on Graph Neural Network (GNN) components that rely on ground-truth variable relationships that are provided in advance or determined by heuristics. However, hardcoded graphs can be difficult to define in domains that do not have clear physical relationships between variables. Even when they do exist, predefined graphs can create fixed (\textit{static}) spatial structure. Variable relationships may change over time and can be more accurately modeled by a context-dependent (\textit{dynamic}) graph. A temporal attention network with spatial graph layers is depicted in Figure \ref{fig:st-graph}c.

This paper proposes a general-purpose multivariate forecaster with the long-term prediction ability of a time series model and the dynamic spatial modeling of a GNN without relying on a hardcoded graph. Let $N$ be the number of variables we are predicting and $L$ be the sequence length in timesteps. We flatten multivariate inputs of shape $(L, N)$ into long sequences of shape $(L \times N, 1)$ where each input token isolates the value of a single variable at a given timestep. The resulting input allows Transformer architectures to learn attention networks across both space and time jointly, creating the ``spatiotemporal attention" mechanism shown in Figure \ref{fig:st-graph}d. Spatiotemporal attention learns dynamic variable relationships purely from data, while an encoder-decoder Transformer architecture enables accurate long-term predictions. Our method avoids TSF/GNN domain knowledge by representing the multivariate forecasting problem in a raw format that relies on end-to-end learning; the cost of that simplicity is the engineering challenge of training Transformers at long sequence lengths. We explore a variety of strategies, including fast attention, hybrid convolutional architectures, local-global and shifted-window attention, and a custom spatiotemporal embedding scheme. Our implementation is efficient enough to run on standard resources and scales well with high-memory GPUs. Extensive experiments demonstrate the benefits of Transformers with spatiotemporal attention in benchmarks from traffic forecasting to electricity production, temperature prediction, and metro ridership. We show that a single approach can achieve highly competitive results against specialized baselines from both the time series forecasting and spatial-temporal GNN literature. We open-source a large codebase that includes our model, datasets, and the groundwork for future research directions.

\section{Background and Related Work}
\label{sec:related}

\subsection{Long-Range Transformers}
\label{sec:related:long-range}
For brevity, we assume the reader is familiar with the Transformer architecture \cite{vaswani2017attention}. A brief overview of the self-attention mechanism can be found in Appendix \ref{app:attention}. This paper will focus on the interpretation of Transformers as a learnable message passing graph amongst a set of inputs \cite{joshi2020transformers, liu2019contextualized, zaheer2020big}. Let $\mathbf{A} \in \mathbb{R}^{L_q \times L_k}$ be the attention scores of a query sequence of length $L_q$ and a key sequence of length $L_k$. $\mathbf{A}$ acts much like the adjacency matrix of a graph between the tokens of the two sequences, where $\mathbf{A}[i, j]$ denotes the strength of the connection between the $i$th token of the query sequence and the $j$th token in the key sequence.

 Because attention involves matching each query to the entire key sequence, its runtime and memory use grows quadratically with the length of its input. As a result, the research community has raced to develop and evaluate Transformer variants for longer sequences \cite{tay2020long}. Many of these methods introduce heuristics to sparsify the attention matrix. For example, we can attend primarily to adjacent input tokens \cite{li2019enhancing}, select global tokens \cite{guo2019star}, increasingly distant tokens \cite{ye2019bp, child2019generating} or a combination thereof \cite{zaheer2020big, zhang2021poolingformer}. While these methods are effective, their inductive biases about the structure of the trained attention matrix are not always compatible with tasks outside of NLP. Another approach looks to approximate attention in sub-quadratic time while retaining its flexibility \cite{wang2020linformer, xiong2021nystromformer, zhu2021longshort, cosformer}. Particularly relevant to this work is the \texttt{Performer} \cite{choromanski2021rethinking}. \texttt{Performer} approximates attention in linear space and time with a kernel of random orthogonal features and enables the long-sequence approach that is central to our work. For a thorough survey of efficient attention mechanisms, see \cite{xformer_survey}.

\subsection{Time Series Forecasting and Transformers}
\label{sec:temporal_related}

Deep learning approaches to TSF are generally based on a seq2seq framework in which a \textit{context window} of the $c$ most recent timesteps is mapped to a \textit{target window} of predictions for a time horizon of $h$ steps into the future. Let $x_t$ be a vector of timestamp values (the day, month, year, etc.) at time $t$ and $y_t$ be a vector of variable values at time $t$. Given a context sequence of time inputs $(x_{T-c}, \dots, x_{T})$ and variables $(y_{T-c}, \dots, y_{T})$ up to time $T$, we output another sequence of variable values $(\hat{y}_{T+1}, \dots, \hat{y}_{T+h})$ corresponding to our predictions at future timesteps $(x_{T+1}, \dots, x_{T+h})$.

The most common class of deep TSF models are based on a combination of Recurrent Neural Networks (RNNs) and one-dimensional convolutions (Conv1Ds) \cite{borovykh2017conditional, smyl2020hybrid, salinas2019deepar, lai2018modeling}. More related to the proposed method is a recent line of work on attention mechanisms that aim to overcome RNNs' autoregressive training and difficulty in interpreting long-term patterns \cite{iwata2020few, li2019enhancing, zerveas2020transformer, wu2020adversarial, oreshkin2020metalearning}. Notable among these is the \texttt{Informer} \cite{zhou2021informer} - a general encoder-decoder Transformer architecture for TSF. \texttt{Informer} takes the sequence timestamps $(x_{T-c}, \dots, x_{T+h})$ and embeds them in a higher dimension $d$. The variable values - with zeros replacing the unknown target sequence $(y_{T-c}, \dots, y_{T}, 0_{T+1} \dots, 0_{T+h})$ - are mapped to the same dimension $d$. The time ($x$) and variable ($y$) components sum to create an input sequence of $c+h$ tokens, written in matrix form as $\mathbf{Z} \in \mathbb{R}^{(c+h) \times d}$. The encoder processes the sub-sequence $\mathbf{Z}[0 \dots c]$ while the decoder observes the target sequence $\mathbf{Z}[c+1 \dots c+h]$. The outputs of the decoder are treated as predictions and error is minimized by regression to the true sequence values. Note that \texttt{Informer} outputs the entire prediction sequence directly in one forward pass, as opposed to decoder-only generative models that output each token iteratively (e.g., large language models). This has the advantage of reducing compute by reusing the encoder representation in each decoder layer and minimizing error accumulation in autoregressive predictions. 

Following \texttt{Informer}, a rapidly expanding line of work has looked to improve upon the benchmark results of Transformers in TSF. These methods adjust the model architecture \cite{yformer, liu2021pyraformer} and re-introduce classic time series domain knowledge such as series decomposition, auto-correlation, or computation in frequency space \cite{fedformer, woo2022etsformer}. These time series biases are often used as a domain-specific form of efficient attention \cite{autoformer, liu2021pyraformer, preformer}, as many long-term TSF tasks naturally extend beyond the sequence limits of default Transformers. It is important to note that time series Transformers use one input token per timestep, so that the embedding of the token at time $t$ represents $N$ distinct variables at that moment in time. This is in contrast with domains like natural language processing in which each token represents just one unified idea (e.g., a single word) \cite{shih2019temporal}. The message-passing graph that results from attention over a multivariate sequence learns patterns across time while keeping variables grouped together (Fig. \ref{fig:st-graph}b). This setup forces the variables within each token to receive the same amount of information from other timesteps, despite the fact that variables may have distinct patterns or relationships to each other. Ideally, we would have a way to model these kinds of variable relationships. 

\subsection{Spatial-Temporal Forecasting and GNNs}
\label{sec:spatiotemporal_related}

Multivariate TSF involves two dimensions of complexity: the forecast sequence's duration, $L$, and the number of variables $N$ considered at each timestep. As $N$ grows, it becomes increasingly important to model the  relationships between each variable. The multivariate forecasting problem can be reformulated as the prediction of a sequence of graphs, where at each timestep $t$ we have a graph $\mathcal{G}_t$ consisting of $N$ variable nodes connected by weighted edges. Given a context sequence of graphs $(\mathcal{G}_{T-c}, \dots, \mathcal{G}_{T})$, we must predict the node values of the target sequence $(\mathcal{G}_{T+1}, \dots, \mathcal{G}_{T+h})$.

Graph Neural Networks (GNNs) are a category of deep learning techniques that aim to explicitly model variable relationships as interactions along a network of nodes \cite{Wu_2021}. Earlier work uses graph convolutional layers \cite{niepert2016learning} to pass information amongst the variables of each timestep while using standard sequence learning architectures (e.g., RNNs \cite{li2018diffusion}, Conv1Ds \cite{yu2018spatio}, dilated Conv1Ds \cite{wu2019graph}) to adapt those representations across time. More recent work has extended this formula by replacing the spatial and/or temporal learning mechanisms with attention modules \cite{cai2020traffic, xu2021spatialtemporal, wu2021traversenet}. Temporal attention with intra-token spatial graph learning is depicted in Figure \ref{fig:st-graph}c. Existing GNN-based methods have a combination of three common shortcomings: 

\begin{enumerate}
    \item Their spatial modules require predefined graphs denoting variable relationships. This can make it more difficult to solve abstract forecasting problems where  relationships are unknown and must be discovered from data.
    \item They perform separate spatial and temporal updates in alternating layers. This creates information bottlenecks that restrict spatiotemporal message passing.
    \item Their spatial graphs remain static across timesteps. Variable relationships can often change over time and should be modeled by a dynamic graph. 
\end{enumerate}

\textbf{Appendix \ref{app:gnn_related} provides a detailed overview of related work in the crowded field of spatial-temporal GNNs, and includes a categorization of existing methods according to these three key differences}. Our goal is to develop a seq2seq time series model with \textit{true spatiotemporal message passing} on a \textit{dynamic graph} that can be competitive with GNNs despite \textit{not using predefined variable relationships}.

\section{Spatiotemporal Transformers}

\begin{figure}[]
    \centering
    \includegraphics[width=\linewidth]{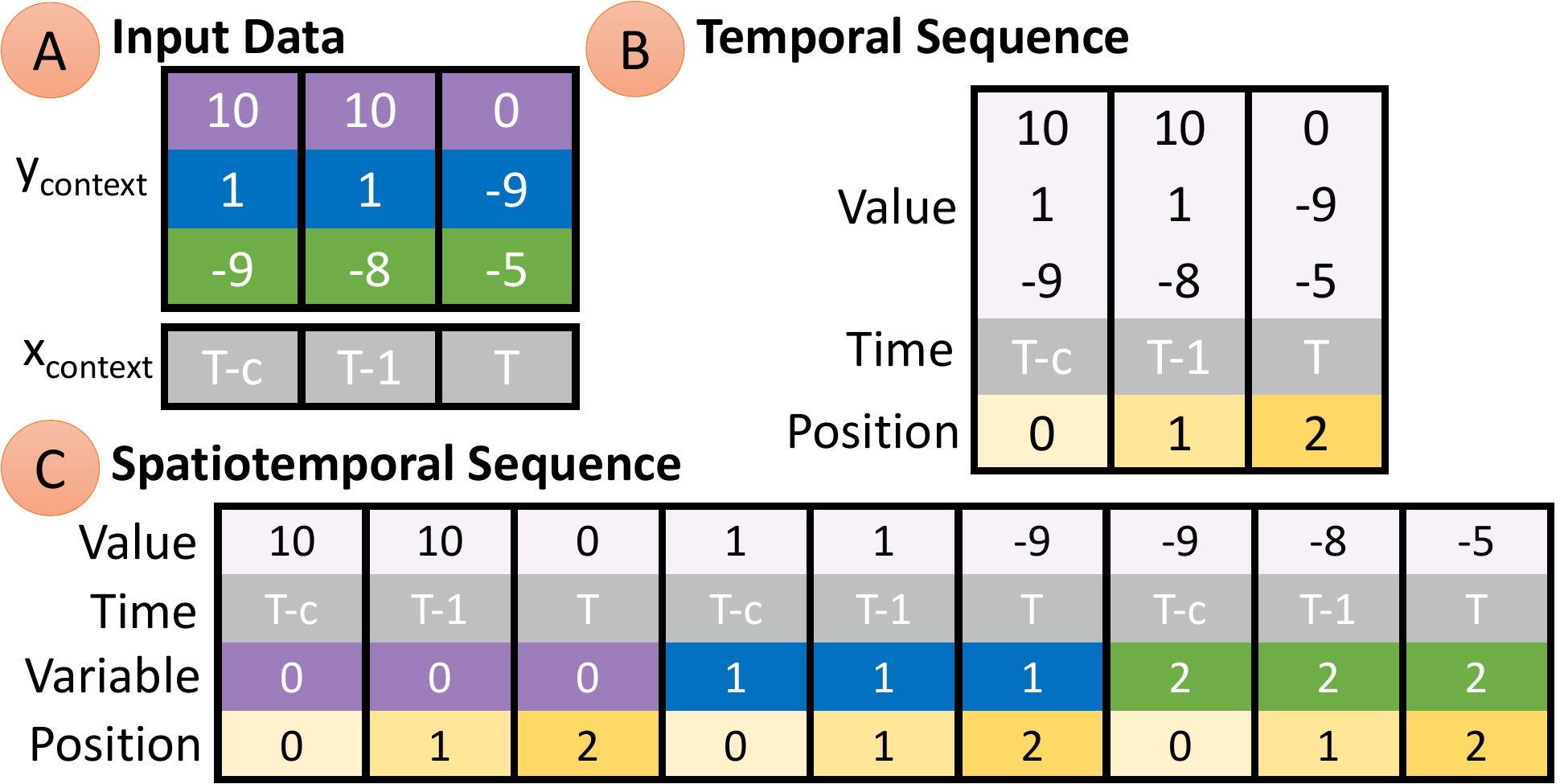}
    \caption{\textbf{Embedding Multivariate Data} (a) The example context sequence from Fig. \ref{fig:st-graph}a. (b) Standard ``Temporal" embedding input sequence where each column will become a token. (c) Flattened spatiotemporal input sequence for Fig. \ref{fig:st-graph}d with position and variable indices.}
    \label{fig:embed_seq}
    \vspace{-9pt}
\end{figure}

\subsection{Spatiotemporal Sequences}
\label{method:intro}
 We begin by building upon the \texttt{Informer}-style encoder-decoder Transformer framework. As discussed in Sec. \ref{sec:temporal_related}, \texttt{Informer} generates $d$-dimensional embeddings of the sequence $\big((x_{T-c}, y_{T-c}), \dots, \\ (x_{T}, y_{T}), (x_{T+1}, 0_{T+1}), \dots, (x_{T+h}, 0_{T+h})\big)$, with the result expressed in matrix form as $\mathbf{Z} \in \mathbb{R}^{(c+h) \times d}$. We propose to modify the token embedding input sequence by flattening each multivariate $y_t$ vector into $N$ scalars with a copy of its timestamp $x_t$, leading to a new sequence: $\big((x_{T-c}, y_{T-c}^{0}), \dots, (x_{T-c}, y_{T-c}^{N}), \dots, (x_{T}, y_{T}^{0}), \dots, \\ (x_{T}, y_{T}^{N}), (x_{T+1}, 0_{T+1}^{0}), \dots, (x_{T+1}, 0_{T+1}^{N}), \dots, (x_{T+h}, 0_{T+h}^{N})\big)$. Embedding this longer sequence results in a $\mathbf{Z'} \in \mathbb{R}^{N(c+h) \times d}$. When we pass $\mathbf{Z'}$ through a Transformer, the attention matrix $\mathbf{A} \in \mathbb{R}^{N(c+h) \times N(c+h)}$ then represents a spatiotemporal graph with a direct path between every variable at every timestep (see Appendix \ref{app:gnn_related} Fig. \ref{fig:st_attn_explained}). In addition, we are now learning spatial relationships that do not rely on predefined variable graphs and can change dynamically according to the time and variable values of the input data. This concept is depicted in Figure \ref{fig:st-graph}d. There are two important questions left to answer:
 
\begin{enumerate}
    \item How do we embed this sequence so that the attention network parameters can accurately interpret the information in each token?
    \item Are we able to multiply the sequence length by a factor of $N$ and still scale to real-world datasets?
\end{enumerate}

\begin{figure}
    \centering
    \includegraphics[width=\columnwidth]{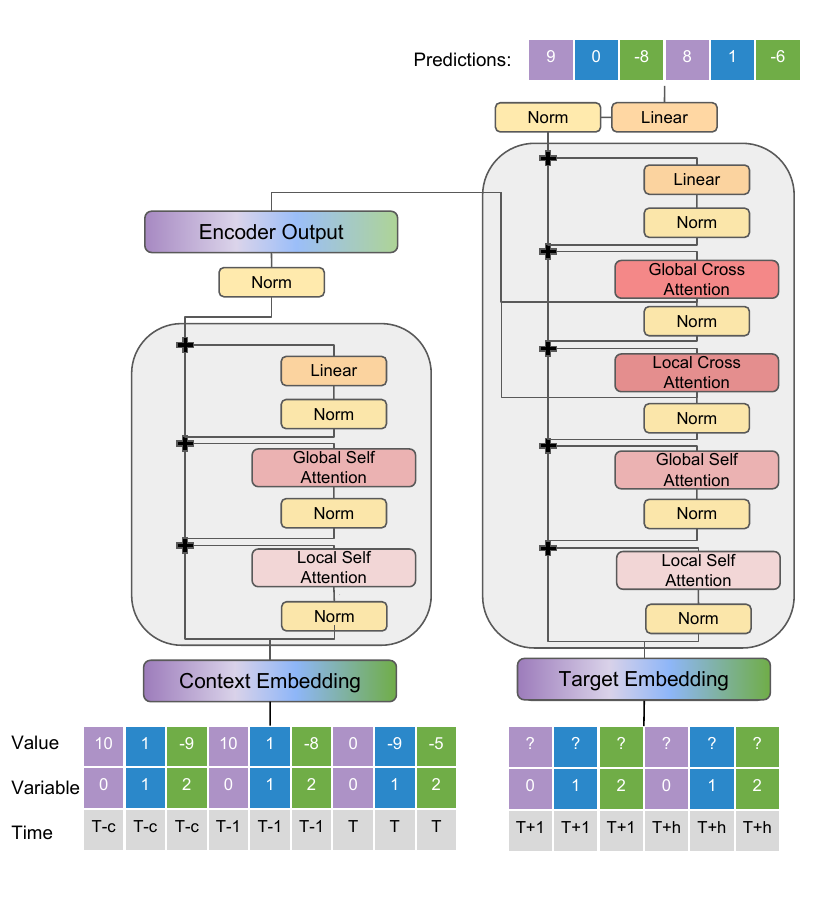}
    \vspace{-20pt}
    \caption{The \method \hspace{1px} architecture for joint spatiotemporal learning applied to the sequence shown in Figure 1a. This architecture creates a practical implementation of the spatiotemporal attention concept in Figure 1d.}
    \label{fig:arch}
\end{figure}

\subsection{Spatiotemporal Embeddings}

\textbf{Representing Time and Value.} The input embedding module of a TSF Transformer determines the way the $(x, y)$ sequence concepts in Sec \ref{method:intro} are implemented in practice. We create input sequences consisting of the values of our time series variables ($y$) and time information ($x$). Although many seq2seq TSF models discard explicit time information, we use \texttt{Time2Vec} layers \cite{kazemi2019time2vec} to learn seasonal characteristics. \texttt{Time2Vec} maps $x_t$ to sinusoidal patterns of learned offsets and wavelengths. This helps represent periodic relationships that extend past the limited length of the context sequence. The concatenated variable values and time embeddings are then projected to the input dimension of the Transformer model with a feed-forward layer. We refer to the resulting output as the ``value+time embedding."

\textbf{Representing Position.} Transformers are permutation invariant, meaning they cannot interpret the order of input tokens by default. This is fixed by adding a \textit{position embedding} to the tokens; we opt for the fully learnable position embedding variant where we initialize $d$-dimensional embedding vectors for each timestep up to the maximum sequence length. An example of the way input time series data is organized for standard TSF Transformers is shown in Figure \ref{fig:embed_seq}a and \ref{fig:embed_seq}b. Existing models use a variety of implementation details, but the end result is a $d$-dimensional token that represents the values of all $N$ variables simultaneously.

\textbf{Representing Space.} We create spatiotemporal sequences by flattening the $N$ variables into separate tokens (Fig. \ref{fig:embed_seq}c). Each token is given a copy of the time information $x$ for its value+time embedding and assigned the position embedding index corresponding to its original timestep order, so that each position now appears $N$ times. We differentiate between the variables at each timestep with an additional ``variable embedding." We initialize $N$ $d$-dimensional embedding vectors for each variable index, much like the position embedding. Note that this means the variable representations are randomly initialized and learned end-to-end during training. The input pattern of position and variable indices is represented by Figure \ref{fig:embed_seq}c. The use of two learnable embedding sets (space and time) creates interesting parallels between our time series model and Transformers in computer vision. Further discussion and small-scale experiments on image data using our model are  in Appendix \ref{app:image_completion}.

\textbf{Representing Missing Data.} Many real-world applications involve missing data. Existing work often ignores timesteps that have any missing variable values (wasting valuable data) or replaces them with an arbitrary scalar that can confuse the model by being  unpredictable or ambiguous. For example, popular traffic benchmarks (Sec. \ref{sec:st_forecasting}) replace missing values with zeros, so that it is unclear whether traffic was low or was not recorded. Embedding each variable in its own separate token gives us the flexibility to leave values missing in the data pipeline, replace them with zeros in the forward pass, and then tell the model when values were originally missing with a binary ``given embedding." The ``value+time", variable, position, and ``given" embeddings sum to create the final spatiotemporal input sequence.

\label{sec:method:st-embed}

\subsection{Scaling to Long Sequence Lengths}
\label{sec:method:scaling}

Our embedding scheme in Sec \ref{sec:method:st-embed} converts a sequence-to-sequence problem of length $L$ into a new format of length $LN$. Standard Transformer layers have a maximum sequence length of less than $1,000$; the spatiotemporal sequence lengths considered in our experimental results exceed $26,000$. Clearly, additional optimization is needed to make these problems feasible.

\textbf{Scaling with Fast-Attention}. However, we are fortunate that there is a rapidly-developing field of research in long-sequence Transformers (Sec. \ref{sec:related:long-range}). The particular choice of attention mechanism is quite flexible, although most results in this paper use \texttt{Performer} FAVOR+ attention \cite{choromanski2021rethinking} - a linear approximation of attention via a random kernel method. The direct (non-iterative) output format of our model does not require causal masking, which can be a key advantage when dealing with approximate attention mechanisms that have difficulty masking an attention matrix that they never explicitly compute. However, there are some domains (Appendix \ref{app:multiseries}) with variable-length sequences and padding that make masking an important consideration.

\textbf{Scaling with Initial and Intermediate Conv1Ds}. In low-resource settings, we can also look for ways to learn shorter representations of the input with strided convolutions. ``Initial" convolutions are applied to the value+time embedding of the encoder. ``Intermediate" convolutions occur between attention layers and are a key component of the \texttt{Informer} architecture. However, our flattened spatiotemporal sequences lay out variables in an arbitrary order (Fig. \ref{fig:embed_seq}c). We rearrange the input so that each variable can be passed through a convolutional block independently and then recombined into the longer spatiotemporal sequence. 

\textbf{Scaling with Shifted Attention Windows}. We can split the input into smaller chunks or ``windows" across time, and perform spatiotemporal attention amongst the tokens of each window separately. Subsequent layers can shift the window boundaries and let information spread between distant windows as the network gets deeper. The shifted window approach is inspired by the connection between Vision Transformers and spatiotemporal attention \cite{swin_transformer} (see Appendix \ref{app:image_completion}).

These scaling methods can be mixed and matched based on available GPU memory. Our goal is to learn spatiotemporal graphs across entire input sequences. Therefore, we try use as few optimizations as possible, even though convolutions and windowed attention have shown promise as a way to improve predictions despite not being a computational necessity. Most results in this paper were collected using fast attention alone with less than $40$GBs of GPU memory. Strided convolutions are only necessary on the largest datasets. Shifted window attention saves meaningful amounts of memory when using quadratic attention, so we primarily use it when we need to mask padded sequences.

\subsection{\method}
\label{sec:arch}
\textbf{Local and Global Architecture}. We find that attention over a longer multivariate sequence can complicate learning in problems with large $N$. We add some architectural bias towards the sequence of each token's own variable with ``local" attention modules in each encoder and decoder layer. In a local layer, each token attends to the timesteps of its own spatial variable. Note that this does not mean we are simplifying the ``global" attention layer by separating temporal and spatial attention, as is common in spatial-temporal methods (Appendix \ref{app:gnn_related}). Rather, tokens attend to every token in their own variable's sequence and then to every token in the entire spatiotemporal global sequence. We use a Pre-Norm architecture \cite{xiong2020layer} and BatchNorm \cite{batchnorm} normalization. Figure \ref{fig:arch} shows a one-layer encoder-decoder architecture.

\textbf{Output and Time Series Tricks}. The final feed-forward layer of the decoder outputs a sequence of predictions that can be folded back into the original input format of ($x_t, \hat{y}_t$). We can then optimize a variety of forecasting loss functions, depending on the particular problem and the baselines we are comparing against. Our goal is to create a general multivariate sequence model, so we try to avoid adding domain specific tricks whenever possible. However, we include features from the recent Transformer TSF literature such as seasonal decomposition, input normalization, and the ability to predict the target sequence as the net difference from the output of a simple linear model. These tricks are turned off by default and only used in non-stationary domains where distribution shift is a major concern; we return to this briefly in the Experiments section and in detail in Appendix \ref{app:ettm1_appendix}. We nickname our full model the \method \hspace{1px} for clarity in experimental results. More implementation details are listed in Appendix \ref{app:implementation}, including explanations of several techniques that are not used in the main experimental results but are part of our open-source code release.

\section{Experiments}

Our experiments are designed to answer the following questions: 
\begin{enumerate}
\item Is our model competitive with seq2seq methods on highly temporal tasks that require long-term forecasting?
\item Is our model competitive with graph-based methods on highly spatial tasks, despite not having access to a predefined variable graph?
\item Does our spatiotemporal sequence formulation let Transformers learn meaningful variable relationships?
\end{enumerate}

We compare against representative methods from the TSF and GNN literature in addition to ablations of our model. \texttt{MTGNN} \cite{wu2020connecting} is a GNN method that is capable of learning its graph structure from data. \texttt{LinearAR} is a basic linear model that iteratively predicts each timestep of the target sequence as a linear combination of the context sequence and previous outputs. We also include a standard encoder-decoder \texttt{LSTM} \cite{lstm} and the RNN/Conv1D-based \texttt{LSTNet} \cite{lai2018modeling}. The most important baseline is an ablation of our method similar to \texttt{Informer} that controls for implementation details to measure the impact of spatiotemporal attention; this model uses the \method \hspace{1mm} architecture with the more standard temporal sequence embedding (see Fig. \ref{fig:embed_seq}b). We refer to this as the \texttt{Temporal} model\footnote{One key detail that cannot be applied to the \texttt{Temporal} model is the use of local attention layers, because there is no concept of local vs. global when tokens represent a combination of variables.}. These baselines are included in our open-source release and use the same training loop and evaluation process. We add \texttt{Time2Vec} information to baseline inputs when applicable because \texttt{Time2Vec} has been shown to improve the performance of a variety of sequence models \cite{kazemi2019time2vec}. Time series forecasting tricks like decomposition and input normalization have been implemented for all methods. We also provide reference results from existing work when they are available, including several spatial-temporal GNN models with predefined graph information (Appendix \ref{app:gnn_related}) and recent TSF Transformers (Sec. \ref{sec:temporal_related}). We report evaluation metrics as the average over all the timesteps of the target sequence, and the average of at least three training runs.

\subsection{Toy Examples} 
\label{sec:toy_experiment}
We begin with a binary multivariate copy task similar to those used to evaluate long-range dependencies in memory-based sequence models \cite{graves2014neural}. However, we add an extra challenge and shift each variable's output by a unique number of timesteps (visualized in Appendix \ref{app:toy_exps} Fig. \ref{fig:shifted_copy}). The shifted copy task was created because it requires each variable to attend to different timesteps; \texttt{Temporal}'s attention is fundamentally unable to do this, and instead resorts to learning one variable relationship per attention head until it runs out of heads and produces blurry outputs. The attention heads are visualized in Appendix \ref{app:toy_exps} Fig. \ref{fig:shifted_copy_temporal_patterns} while an example sequence result is shown in Appendix \ref{app:toy_exps} Fig. \ref{fig:shifted_copy_temporal}. Spatiotemporal attention is capable of learning all $N$ variable relationships in one attention head (Appendix \ref{app:toy_exps} Fig. \ref{fig:shifted_copy_spatiotemporal_patterns}), leading to an accurate output (Appendix \ref{app:toy_exps} Fig. \ref{fig:shifted_copy_spatiotemporal}). 

Next we look at a more traditional forecasting setup inspired by \cite{shih2019temporal} consisting of a multivariate sine-wave sequence with strong inter-variable dependence. Several ablations of our method are considered. This dataset creates a less extreme example of the effect in the shifted copy task, where \texttt{Temporal} models are forced to compromise their attention over timesteps in a way that reduces predictive power over variables with such distinct frequencies. Our method learns an uncompromising spatiotemporal relationship among all tokens to generate the most accurate predictions. Dataset details and results can be found in Appendix \ref{app:toy_exps}.

\subsection{Time Series Forecasting}

\textbf{NY-TX Weather.} We continue with more realistic time series domains where we must learn to forecast a relatively small number of variables $N$ with unknown relationships over a long duration $L$. First we evaluate on a custom dataset of temperature values compiled from the ASOS Weather Network \cite{asos}. We use three stations in central Texas and three in eastern New York to create two almost unrelated spatial sub-graphs. Temperature values are taken at one-hour intervals, and we investigate the impact of sequence length by predicting $40$, $80$, and $160$ hour target sequences. The results are shown in Table \ref{tbl:asos_results}. Our spatiotemporal embedding scheme provides the most accurate forecasts, and its improvement over the \texttt{Temporal} method appears to increase over longer sequences where the lack of flexibility that comes from grouping the two geographic regions together may become more relevant. \texttt{MTGNN} learns spatial relationships, but temporal consistency can be difficult without decoder attention; its convolution-only output mechanism begins to struggle at the $80$ and $160$ hour lengths.

\begin{table}[]
\resizebox{\columnwidth}{!}{
\begin{tabular}{@{}lccccc@{}}
\toprule
                  & \textbf{LinearAR} & \textbf{LSTM} & \textbf{MTGNN} & \textbf{Temporal} & \method  \\ \midrule
\textbf{40 hours} &                 &               &                &                   &                         \\ \midrule
MSE      & 18.84           & 14.29         & 13.32          & 13.29             & {\underline{12.49}}             \\
MAE               & 3.24            & 2.84          & 2.67           & 2.67              & {\underline{2.57}}              \\
RRSE              & 0.40             & 0.355         & 0.34           & 0.34              & {\underline{0.33}}              \\ \midrule
\textbf{80 hours} &                 &               &                &                   &                         \\ \midrule
MSE      & 23.99           & 18.75         & 19.27          & 19.99             & {\underline{17.9}}              \\
MAE               & 3.72            & 3.29          & 3.31           & 3.37              & {\underline{3.19}}              \\
RRSE              & 0.45            & {\underline{0.40}}     & 0.41           & 0.41              & {\underline{0.40}}               \\ \midrule
\textbf{160 hours} &                 &               &                &                   &                         \\ \midrule
MSE      & 28.84           & 22.11         & 24.28          & 24.16             & {\underline{21.35}}                 \\
MAE               & 4.13            & 3.63          & 3.78           & 3.77              & {\underline{3.51}}                 \\
RRSE              & 0.50             & {\underline{0.44}}          & 0.46           & 0.46              & {\underline{0.44}}                 \\ \bottomrule
\end{tabular}
}
\caption{\textbf{NY-TX Weather Results.}}
\label{tbl:asos_results}
\vspace{-10pt}
\end{table}

\textbf{ETTm1.} The Transformer TSF literature (Sec. \ref{sec:temporal_related}) has settled on a set of common benchmarks for experimental comparison in long-term forecasting, including a dataset of electricity transformer temperature (ETT) series introduced by \cite{zhou2021informer}. We compare against \texttt{Informer} and a selection of follow-up methods on the multivariate minute resolution variant in Table \ref{tbl:ettm1}. At first glance, \method\hspace{1mm} offers meaningful improvements to \texttt{Informer} and is competitive with later variants that make use of additional time series domain knowledge. However, our work on ETTm1 and similar benchmarks revealed that these results have less to do with advanced attention mechanisms or model architectures than they do with robustness to the distribution shift caused by non-stationary datasets. This is highlighted by the performance of simple linear models like \texttt{LinearAR}, which closely matches the accuracy of \texttt{ETSFormer}. In fact, a few tricks allow most models to achieve near state-of-the-art performance; reversible instance normalization \cite{kim2021reversible}, for example, is enough to more than halve the prediction error of the \texttt{LSTM} baseline - noticeably outperforming the original \texttt{Informer} results. This discussion is continued in detail in Appendix \ref{app:ettm1_appendix} with experiments on ETTm1 and another common benchmark. In addition, we create a custom task to investigate the way time series models handle different kinds of distributional shift.

\begin{table}[]
\resizebox{\columnwidth}{!}{
\begin{tabular}{cccccc}
\hline
                                   & \multicolumn{5}{c}{Prediction Length}                                                                                       \\
                                   & 24                          & 48                          & 96                          & 288                         & 672                         \\ \hline
LSTM                      & 0.63                        & 0.94                        & 0.91                        & 1.12                        & 1.56                        \\
Informer                   & 0.37                        & 0.50                        & 0.61                        & 0.79                        & 0.93                        \\ \hline
Pyraformer                         &                             &                             & 0.49                        & 0.66                        & 0.71                        \\
YFormer                           & 0.36                        & 0.46                        & 0.57                        & 0.59                        & 0.66                        \\
Preformer                          & 0.40                        & 0.43                        & 0.45                        & 0.49                        & 0.54                        \\
Autoformer                         & 0.40                        & 0.45                        & 0.46                        & 0.53                        & 0.54                        \\
ETSFormer                & 0.34                        & 0.38                        & \underline{0.39}                        & {\underline{0.42}}                  & {\underline{0.45}}   \\

LinearAR & \underline{0.33}                        & \underline{0.37}                  & \underline{0.39}                  & 0.44                        & 0.48

\\ \hline
\method  & 0.34                        & 0.38                        & 0.40                        & 0.45                        & 0.52                        \\ \hline
\end{tabular}
}
\caption{\textbf{ETTm1 test set normalized MAE.} Additional results and discussion provided in Appendix \ref{app:ettm1_appendix}.}
\label{tbl:ettm1}
\vspace{-10pt}
\end{table}

\subsection{Spatial-Temporal Forecasting}
\label{sec:st_forecasting}

\textbf{AL Solar.} We turn our focus to problems on the GNN end of the spatiotemporal spectrum where $N$ approaches or exceeds $L$. The AL Solar dataset consists of solar power production measurements taken at $10$-minute intervals from $137$ locations. We predict $4$-hour horizons, leading to the longest spatiotemporal sequences of our main experiments; the results are shown in Table \ref{tbl:al_solar_results}. \method \hspace{1px} is significantly more accurate than the TSF baselines. We speculate that this is due to an increased ability to forecast unusual changes in power production due to weather or other localized effects. \texttt{MTGNN} learns similar spatial relationships, but its temporal predictions are not as accurate.

\begin{table}[h]
\resizebox{\columnwidth}{!}{
\begin{tabular}{@{}lcccccc@{}}
\toprule
             & \textbf{LinearAR} & \textbf{LSTNet} & \textbf{LSTM} & \textbf{MTGNN} & \textbf{Temporal} & \method \\ \midrule
MSE & 14.3            & 15.09           & 10.59          & 11.40          & 9.94              & \underline{7.75}        \\
MAE          & 2.29            & 2.08            & 1.56          & 1.76           & 1.60              & \underline{1.37}        \\ \bottomrule
\end{tabular}
}
\caption{\textbf{AL Solar Results.}}
\label{tbl:al_solar_results}
\vspace{-10mm}
\end{table}

\begin{table*}[]
\resizebox{\linewidth}{!}{
\begin{tabular}{cccccccccc|c}
\hline
\multicolumn{1}{l|}{} & \multicolumn{3}{c|}{\textbf{Time Series Models}} & \multicolumn{6}{c|}{\textbf{ST-GNN Models}} &  \\ \hline
\multicolumn{1}{l|}{} & LinearAR & LSTM & \multicolumn{1}{c|}{Temporal} & \textit{DCRNN} & \textit{MTGNN} & \textit{STAWnet} & \textit{ST-GRAT} & \textit{\begin{tabular}[c]{@{}c@{}}Graph\\ WaveNet\end{tabular}} & \multicolumn{1}{l|}{\textit{\begin{tabular}[c]{@{}l@{}}Traffic\\ Trans.\end{tabular}}} & \method \\ \hline
\multicolumn{1}{l}{\textbf{Metr-LA}} & \multicolumn{1}{l}{} & \multicolumn{1}{l}{} & \multicolumn{1}{l}{} & \multicolumn{1}{l}{} & \multicolumn{1}{l}{} & \multicolumn{1}{l}{} & \multicolumn{1}{l}{} & \multicolumn{1}{l}{} & \multicolumn{1}{l|}{} & \multicolumn{1}{l}{} \\ \hline
\multicolumn{1}{c|}{MAE} & 4.71 & 3.87 & \multicolumn{1}{c|}{3.59} & 3.03 & 3.08 & 3.06 & 3.03 & 3.09 & {\underline{2.83}} & 2.86 \\
\multicolumn{1}{c|}{MSE} & 94.11 & 47.5 & \multicolumn{1}{c|}{52.73} & 37.88 & 39.05 & 38.73 & 39.77 & 39.84 & {\underline{32.82}} & 38.27 \\
\multicolumn{1}{c|}{MAPE} & 12.7 & 10.7 & \multicolumn{1}{c|}{10.7} & 8.27 & 8.30 & 8.34 & 8.25 & 8.42 & {\underline{7.70}} & 7.80 \\ \hline
\multicolumn{1}{l}{\textbf{Pems-Bay}} &  &  &  &  &  &  &  &  &  &  \\ \hline
\multicolumn{1}{c|}{MAE} & 2.24 & 2.41 & \multicolumn{1}{c|}{2.49} & 1.59 & 1.64 & 1.61 & 1.62 & 1.63 & {\underline{1.53}} & 1.61 \\
\multicolumn{1}{c|}{MSE} & 27.62 & 25.49 & \multicolumn{1}{c|}{27.27} & 13.69 & 13.98 & 13.47 & 13.85 & 13.87 & {\underline{13.25}} & 13.99 \\
\multicolumn{1}{c|}{MAPE} & 4.98 & 5.81 & \multicolumn{1}{c|}{6.12} & 3.61 & 3.66 & 3.63 & 3.65 & 3.67 & {\underline{3.49}} & 3.63 \\ \hline
\end{tabular}
}
\caption{\textbf{Traffic Forecasting Results.} Results for italicized models taken directly from published work.}
\label{tbl:traffic_results}
\end{table*}

\begin{table*}[thb]
\begin{tabular}{@{}lcccccccc@{}}
\toprule
 & LSTM & Temporal & \textit{ASTGCN} & \textit{DCRNN} & \textit{GCRNN} & \textit{Graph-WaveNet} & \textit{PVGCN} & \method \\ \midrule
MAE & 29.1 & 29.0 & 28.0 & 26.1 & 26.1 & 26.5 & \underline{23.8} & 25.7 $\pm$ .3 \\
RMSE & 51.3 & 48.5 & 47.2 & 44.64 & 44.5 & 44.8 & \underline{40.1} & 44.7 $\pm$ 1.6 \\ \bottomrule
\end{tabular}
\caption{\textbf{HZMetro Ridership Prediction Results.} Italicized model results provided by \cite{liu2020physical}.}
\label{tbl:metro_results}
\vspace{-10pt}
\end{table*}

\textbf{Traffic Prediction.} Next, we experiment on two datasets common in GNN research. The Metr-LA and Pems-Bay datasets consist of traffic speed readings at $5$ minute intervals, and we forecast the conditions for the next hour. For these experiments we include results directly from the literature (Appendix \ref{app:gnn_related}) to get a better comparison with GNN-based spatial models that used predefined road graphs. The results are listed in Table \ref{tbl:traffic_results}. Our method clearly separates itself from TSF models and enters the performance band of dedicated GNN methods on both datasets (without needing predefined graphs).

\textbf{HZMetro.} \cite{liu2020physical} experiment with  a dataset of passenger arrivals and departures at metro stations in Hangzhou, China for a total of $160$ variables recorded in $15$ minute time intervals. We forecast the next hour and compare against their published results in Table \ref{tbl:metro_results}. \texttt{Spacetimeformer} again shows that it can be as effective as graph-based methods in spatial domains without requiring predefined graphs. We list the bounds of several trials because the performance gap between methods on this dataset is relatively slim.

\begin{figure}[h!]
    \centering
    \includegraphics[width=\columnwidth]{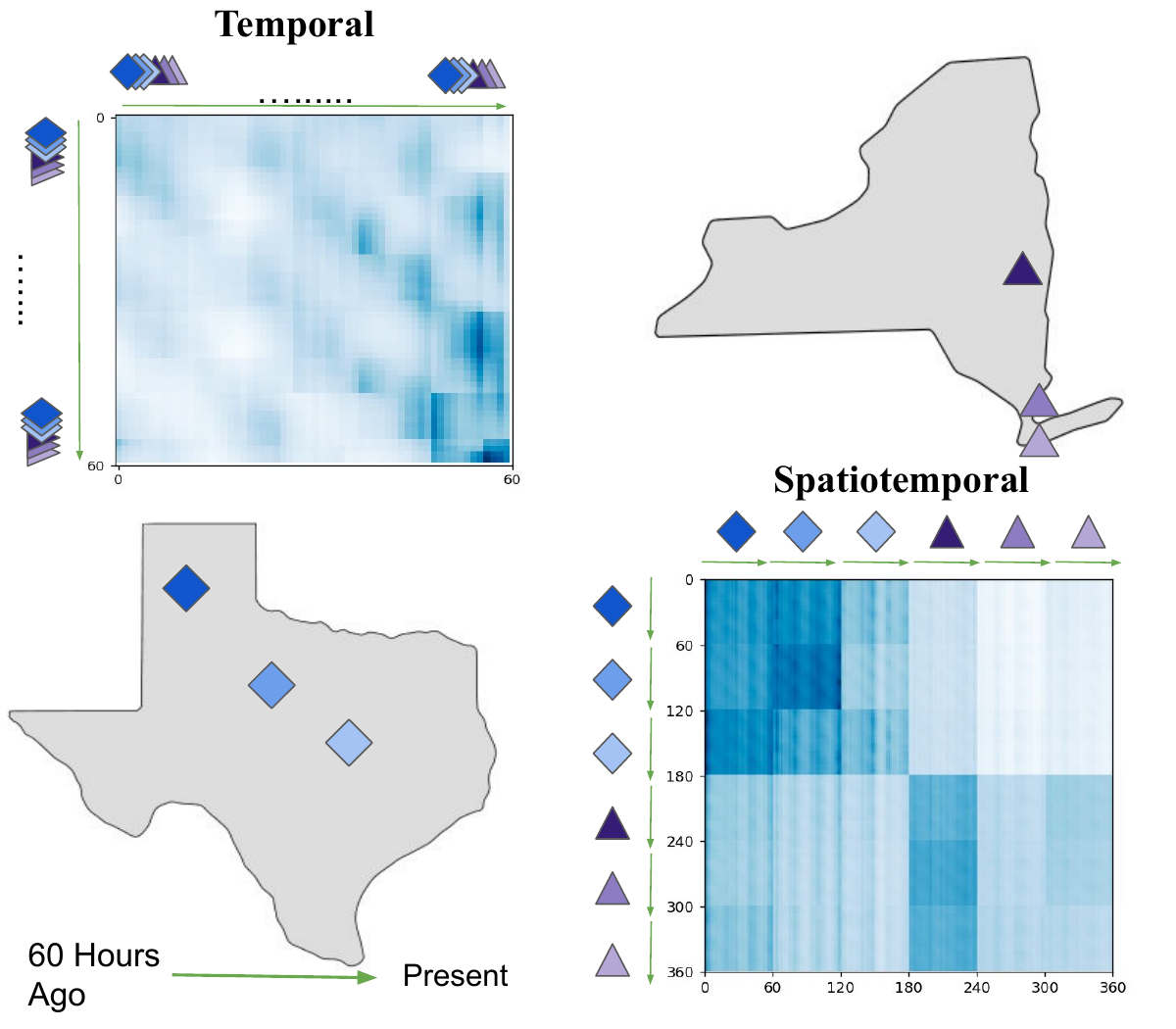}
    \caption{\textbf{Discovering Spatial Relationships from Data:} We forecast the temperature at three weather stations in Texas (lower left, blue diamonds) and three stations in New York (upper right, purple triangles). Temporal attention stacks all six time series into one input variable and attends across time alone (upper left). Our method recovers the correct spatial relationship between the variables along with the strided temporal relation (lower right) (Dark blue shaded entries $\rightarrow$ more attention).}
    \label{fig:asos-attn}
    \vspace{-5mm}
\end{figure}

\textbf{Spatiotemporal Attention Patterns.} Standard Transformers learn sliding attention patterns that resemble convolutions. Our method learns distinct connections between variables - this leads to attention diagrams that tend to be structured in ``variable blocks" due to the way we flatten our input sequence (Fig. \ref{fig:embed_seq}c). Figure \ref{fig:asos-attn} provides an annotated example for the NY-TX weather dataset. Some attention heads convincingly recover the ground-truth relationship between input variables. Our method's GNN-level performance in highly spatial tasks like traffic forecasting supports a similar conclusion when facing more complex graphs.

\textbf{Ablations and Node Classification.} Finally, we perform several ablation experiments to measure the importance of design decisions in our embedding mechanism and model architecture using the NY-TX Weather and Metr-LA Traffic datasets. Results and analysis can be found in Appendix \ref{app:ablations}.

\section{Conclusion and Future Directions}

This paper has presented a unified method for multivariate forecasting based on the application of a custom long-range Transformer architecture to elongated spatiotemporal input sequences. Our approach jointly learns temporal and spatial relationships to achieve competitive results on long-sequence time series forecasting, and can scale to high dimensional spatial problems without relying on a predefined graph. We see several promising directions for future development. First, there is room to scale \texttt{Spacetimeformer} to much larger domains and model sizes. This could be accomplished with additional computational resources or by making better use of optimizations like windowed attention and convolutional layers that were underutilized in our experimental results. Next, our main experiments focus on established benchmarks with relatively static spatial relationships and a standard multivariate input format. While it was necessary to verify that our model is competitive in popular domains like traffic forecasting, we feel that new applications with more rapid changes in variable behavior could take better advantage our fully dynamic and learnable graph. Our embedding scheme also enables flexible input formats with irregularly/unevenly sampled series and exogenous variables (Appendix \ref{app:implementation}). Finally, we see an opportunity to experiment with multi-dataset generalization as is common for Transformers in many other areas of machine learning. Appendix \ref{app:multiseries} provides further discussion of this direction, and the foundation for this work is included in our open-source release.

\printbibliography

\appendix

\section{Additional Background and Related Work}

\subsection{Transformers and Self Attention}
\label{app:attention}

The Transformer \cite{vaswani2017attention} is a deep learning architecture for sequence-to-sequence prediction that is widely used in natural language processing (NLP) \cite{devlin2019bert}. Transformers operate on two sequences of $d$-dimensional vectors, represented in matrix form as $\mathbf{X} \in \mathbb{R}^{L_x \times d}$ and $\mathbf{Z} \in \mathbb{R}^{L_z \times d}$, where $L_x$ and $L_z$ are sequence lengths. The primary component of the model is the attention mechanism that updates the representation of tokens in $\mathbf{X}$ with information from $\mathbf{Z}$. Tokens in $\mathbf{Z}$ are mapped to \textit{key} vectors with learned parameters $W^K$, while tokens in $\mathbf{X}$ generate \textit{query} vectors with $W^Q$. The dot-product similarity between query and key vectors is re-normalized to determine the attention matrix $A(\mathbf{X}, \mathbf{Z}) \in \mathbb{R}^{L_x \times L_z}$:

\begin{align}
    A(\mathbf{X}, \mathbf{Z}) = \text{softmax}\left(\frac{W^Q(\mathbf{X})(W^K(\mathbf{Z}))^T}{\sqrt{d}}\right)
    \label{eq:attnmatrix}
\end{align}

As mentioned in the main text, our method focuses on the interpretation of attention as a form of message passing along a dynamically generated adjacency matrix \cite{joshi2020transformers, liu2019contextualized, zaheer2020big}, where $A(\mathbf{X}, \mathbf{Z})[i, j]$ denotes the strength of the connection between the $i$th token in $\mathbf{X}$ and the $j$th token in $\mathbf{Z}$. The information passed along the edges of this graph are \textit{value} vectors of sequence $\mathbf{Z}$, generated with parameters $W^V$. We create a new representation of sequence $\mathbf{X}$ according to the attention-weighted sum of $W^V(\mathbf{Z})$:

\begin{align}
    \text{Attention}(\mathbf{X}, \mathbf{Z}) \in \mathbb{R}^{L_x \times d} := A(\mathbf{X}, \mathbf{Z})W^V(\mathbf{Z})
    \label{eq:attneq}
\end{align}

Let $\mathbf{Z}$ correspond to the context sequence of the time series forecasting framework (Sec. \ref{sec:temporal_related}) and $\mathbf{X}$ be the target sequence. In an encoder-decoder Transformer, a stack of consecutive encoder layers observe the context sequence and perform self-attention between that sequence and itself ($\text{Attention}(\mathbf{Z}, \mathbf{Z})$), leading to an updated representation of the context sequence $\mathbf{Z'}$. Decoder layers process the target sequence $\mathbf{X}$ and alternate between self-attention ($\text{Attention}(\mathbf{X}, \mathbf{X})$) and cross-attention with the output of the encoder ($\text{Attention}(\mathbf{X}, \mathbf{Z'})$). Each encoder and decoder layer also includes normalization, fully connected layers, and residual connections applied to each token \cite{vaswani2017attention, xiong2020layer}. The output of the last decoder layer is passed through a linear transformation to generate a sequence of predictions.

\subsection{Spatial-Temporal Graph Neural Networks}
\label{app:gnn_related}

Here we expand upon the discussion of spatial-temporal related work summarized by Section \ref{sec:spatiotemporal_related}. The method proposed in this paper is focused on ``spatiotemporal" forecasting, where we learn both spatial relationships amongst multiple variables and temporal relationships across multiple timesteps. There are countless papers on sequence models for forecasting that learn representations across time, and a growing literature on Graph Neural Network (GNN) models that learn representations across space. At the intersection of seq2seq TSF and GNNs  is the spatial-temporal GNN (ST-GNN) literature centered around graph operations over a sequence of variable nodes. The spatial-temporal literature can be difficult to categorize due to heavy overloading of vocabulary like ``spatial attention" and concurrent publication of similar methods. In this section, we try to abstract away implementation details and categorize existing works based on the ways they learn representations of multivariate data.

We begin with non-GNN seq2seq time series models as a point of reference. \texttt{Informer}, for example, groups $N$ variables per timestep into $d$ dimensional vector representations, and then uses self attention to share information between timesteps (Sec. \ref{sec:temporal_related}). This makes models like \texttt{Informer} and later variants (\texttt{Autoformer}, \texttt{ETSFormer}, etc.) purely temporal methods. The Temporal category is not specific to self attention but to any method that learns patterns across time. For example, the LSTM model used in our experiments learns temporal patterns by selectively updating a compressed representation of past timesteps given the current timestep.

ST-GNNs reformulate multivariate forecasting as the prediction of a sequence of graphs. A graph at timestep $t$, denoted $\mathcal{G}_t$, has a set of $N$ nodes ($\mathcal{V}_t$) connected by edges ($\mathcal{E}_t$). The $i$th node $v_t^i$ contains a vector of values/attributes $y_t^i$, with the attributes of all $N$ nodes represented by a matrix $Y_t$. Nodes are connected by weighted edges that can be represented by an adjacency matrix $\mathbf{A}_t \in \mathbb{R}^{N \times N}$. In traffic forecasting, for example, nodes are road sensors with the current traffic velocity as attributes and weighted edges corresponding to road lengths. Using graph notation, the forecasting problem can be written as the prediction of future node values $[Y_{T+1}, Y_{T+2}, \dots, Y_{T+h}]$ given previous graphs $[\mathcal{G}_{T-c}, \dots, \mathcal{G}_T]$, where $h$ is the horizon and $c$ is the context length. In practice, we typically look to learn a parameterized function $f_{\theta}$ from past node values and a fixed (\textit{static}) adjacency matrix:
            
            \begin{align}
                [Y_{T+1}, Y_{T+2}, \dots, Y_{T+h}] = f_{\theta}([Y_{T-c}, \dots, Y_T; \mathbf{A}])
                \label{eq:graph_static_eq}
            \end{align}

\texttt{Traffic Transformer} \cite{cai2020traffic} is a prototypical example of using the graph-based formulation with self-attention components. \texttt{Traffic Transformer} uses the predefined adjacency matrix $\mathbf{A}$ to perform a graph convolution, sharing information between the nodes of each timestep according to the hard-coded spatial relationships. The node representations are then passed through a more standard Transformer component that handles temporal learning. The overall architecture is one example of the ``spatial, then temporal" pattern that is common in ST-GNNs, and is depicted in Figure \ref{fig:st-graph}c.  Once again, this framework does not require an attention mechanism that looks like a Transformer - it just needs a way to share information across space and time in an alternating fashion. \texttt{DCRNN}, for example, is a foundational work in this literature that merges diffusion based graph layers with the recurrent temporal processing of an RNN.

One drawback of graph convolutions on hardcoded adjacency matrices is their inability to adapt spatial connections for specific points in time. In our traffic forecasting example, roadway connections may change due to closure, accidents, or high volume. More formally, some domains require an extension of Eq. \ref{eq:graph_static_eq} to a \textit{dynamic} sequence of adjacency matrices:

 \begin{align}
                [Y_{T+1}, \dots, Y_{T+h}] = f_{\theta}([(Y_{T-c}, \mathbf{A}_{T-c}), \dots, (Y_T, \mathbf{A}_T)])
                \label{eq:graph_dynamic_eq}
\end{align}

However it is rare to have such a sequence available, so practical implementations often rely on attention mechanisms to re-weight their spatial relationships based on time and node attributes, as in \texttt{ASTGCN} \cite{astgcn}. This can allow for some dynamic adaptation of the spatial graph, although we are still unable to learn edges that were not provided in advance. \texttt{GMAN} \cite{zheng2020gman} enables a fully dynamic spatial graph by using $\mathbf{A}$ to create a spatiotemporal embedding for each node $v \in \mathcal{V}_t$. Spatial attention modules based on these embeddings combine with more typical temporal self attention for accurate predictions. 

While adjacency matrices may be available for traffic forecasting where road networks are clearly defined, many multivariate domains have unknown spatial relationships that need to be discovered from data. One approach - used by \texttt{Graph WaveNet} \cite{wu2019graph}, \texttt{MTGNN} \cite{wu2020connecting}, \texttt{AGCRN} \cite{bai2020adaptive}, and others - is to randomly initialize trainable node embeddings and use the similarity scores between them to construct a learned adjacency matrix. However, these graphs are updated by gradient descent and are then static after training is complete. Methods like \texttt{STAWnet} \cite{tian2021spatial} learn dynamic graphs by making the spatial relationships dependent on both the node embeddings and the time/value of the current input. 

\begin{figure}[h!]
    \centering
    \includegraphics{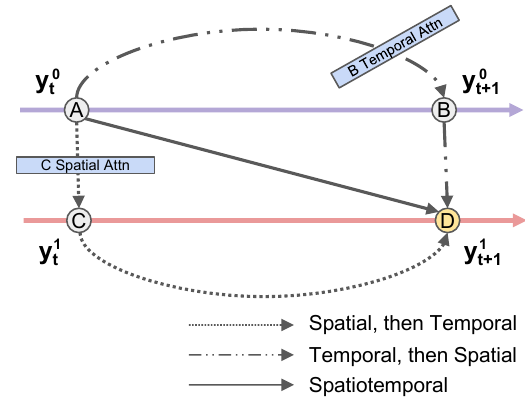}
    \caption{\textbf{“Spatial and Temporal" vs. Spatiotemporal Attention.} We depict a two variable series with purple (top) and red (bottom) variable sequences. Nodes corresponding to timesteps $t$ and $t+1$ and have been labeled A, B, C, D for simplicity. Suppose that the information at node A is important to the representation we want to produce at node D. Alternating temporal and spatial layers force this information to take an indirect route through the attention mechanism of an unrelated token (e.g., B or C). The A $\rightarrow$ D path is then dependent on the similarity of B and A or C and A, as well as any other tokens involved in B’s temporal attention or C’s spatial attention. In contrast, true spatiotemporal attention creates a direct path A $\rightarrow$ D with no risk of information loss.}
    \label{fig:st_attn_explained}
\end{figure}

More recent work takes spatial-temporal learning a step further. Rather than alternating between spatial and temporal layers, fully spatiotemporal methods spread information across space and time jointly by adding edges between the graphs of multiple timesteps. When using alternating spatial/temporal layers, information from past timesteps of neighboring nodes must take an indirect route through the representation of another node. In other words, a spatial layer must store irrelevant information in a node just so that it can be moved by a temporal layer to the timestep where it is relevant and vice versa. This effect is most evident in attention-based models where two-step message passing relies on the queries/keys of unrelated nodes, but also occurs in recurrent/convolutional models due to the need to compress information into a fixed amount of space. A illustrative example using self-attention terminology is provided in Figure \ref{fig:st_attn_explained}. \texttt{STSGCN} \cite{stsgcn}, \texttt{STJGCN} \cite{zheng2021spatiotemporal}, and \texttt{TraverseNet} \cite{wu2021traversenet} expand their predefined graphs by connecting node neighborhoods across short segments of time. \method \hspace{1mm} processes joint spatiotemporal relationships across a learnable dynamic graph, providing the most flexible and assumption-free combination of the ST-GNN literature. We accomplish this by leveraging efficient attention mechanisms to dynamically generate the adjacency matrix of a densely connected graph between all nodes at all timesteps.

In summary, the ST-GNN literature has three key methodological differences that define models' flexibility and accuracy:
\begin{enumerate}
    \item The type of graph used in learning. Are we performing spatial and temporal updates in an alternating fashion, or can we learn relationships across space and time jointly with a true spatiotemporal graph?
    \item The requirement of a predefined adjacency matrix based on known relationships between nodes.
    \item The ability to dynamically adapt spatial relationships according to the current timestep and node values.
\end{enumerate}
\textbf{We categorize related work according to these characteristics in Table \ref{tbl:related_work}}. For more background on ST-GNNs, see \cite{bui2021spatial}.

\begin{table*}
\centering

\begin{tabular}{@{}lcccc@{}}
\toprule
\multicolumn{1}{c}{Method} &  \begin{tabular}[c]{@{}c@{}}Message Passing \\ Type\end{tabular} & \begin{tabular}[c]{@{}c@{}}Predefined \\ Spatial Graph \\ \end{tabular} & \begin{tabular}[c]{@{}c@{}} Dynamic \\ Spatial Graph\end{tabular} \\ \midrule
Informer \cite{zhou2021informer}                   & Temporal              & \xmark & \xmark           \\
TFT \cite{lim2020temporal}                        & Temporal             & \xmark    & \xmark      \\ \midrule

DCRNN \cite{li2018diffusion} & Spatial + Temporal & \checkmark & \xmark \\
TSE-SC \cite{cai2020traffic}             & Spatial + Temporal            & \checkmark  & \xmark \\
PVCGN \cite{liu2020physical} & Spatial + Temporal & \checkmark & \xmark \\
Multi-STGCnet \cite{ye2020multi} & Spatial + Temporal & \checkmark & \xmark \\
STFGNN \cite{stfgnn}                       & Spatial + Temporal & \checkmark & \xmark \\

ASTGCN \cite{astgcn} & Spatial + Temporal & \checkmark & \checkmark- \\
ST-GRAT \cite{Park_2020}                    & Spatial + Temporal            & \checkmark      & \checkmark-  \\
DMGCRN \cite{qin2021dmgcrn} & Spatial + Temporal & \checkmark & \checkmark- \\
STWave \cite{fang2021spatio} & Spatial + Temporal & \checkmark & \checkmark- \\
MGT \cite{mgt2022} & Spatial + Temporal & \checkmark & \checkmark- \\

STTN \cite{xu2021spatialtemporal}                       & Spatial + Temporal            & \checkmark   & \checkmark \\
STIDGCN \cite{stidgcn} & Spatial + Temporal & \checkmark & \checkmark \\
GMAN \cite{zheng2020gman} & Spatial + Temporal & \checkmark & \checkmark \\
STNN \cite{yin2021stnn} & Spatial + Temporal & \checkmark & \checkmark \\
\midrule

Graph WaveNet \cite{wu2019graph} & Spatial + Temporal & \xmark & \xmark \\
MTGNN \cite{wu2020connecting}                      & Spatial + Temporal            & \xmark    & \xmark \\
AGCRN \cite{bai2020adaptive} & Spatial + Temporal & \xmark & \xmark \\
DSTAGNN \cite{dstagnn} & Spatial + Temporal & \xmark & \checkmark- \\
ST-WA \cite{cirstea2022towards} & Spatial + Temporal & \xmark & \checkmark \\
TVGCN \cite{wang2022tvgcn} & Spatial + Temporal & \xmark & \checkmark \\
STAAN \cite{weikang2022spatial} & Spatial + Temporal & \xmark & \checkmark \\
TPA-LSTM \cite{shih2019temporal}                   & Spatial + Temporal            & \xmark     & \checkmark \\
STAWNet \cite{tian2021spatial} & Spatial + Temporal & \xmark & \checkmark \\
STAM \cite{gangopadhyay2021spatiotemporal}                       & Spatial + Temporal            & \xmark   & \checkmark   \\    \midrule

STSGCN \cite{stsgcn} & Short-Term Spatiotemporal & \checkmark  & \xmark \\
TraverseNet \cite{wu2021traversenet}                & Short-Term Spatiotemporal             & \checkmark   & \checkmark-  \\
STJGCN \cite{zheng2021spatiotemporal}               & Short-Term Spatiotemporal             & \checkmark  & \checkmark      \\ \midrule
\textbf{Spacetimeformer}   & \textbf{Spatiotemporal}    & \textbf{\xmark}    & \checkmark \\ \bottomrule
\end{tabular}

\caption{Spatial-Temporal forecasting related work categorized by graph type (Fig \ref{fig:st_attn_explained}), the requirement of a hard-coded variable graph, and the ability to dynamically adapt spatial relationships across time. ``Short-Term Spatiotemporal" refers to spatiotemporal graphs that are restricted to a short range of timesteps. The ``\checkmark-" rating in the dynamic spatial graph column indicates models that re-weight their adjacency matrix without creating new connections.}
\label{tbl:related_work}
\end{table*}

\subsection{Connection to Vision Transformers}
\label{app:image_completion}
\begin{figure}
    \centering
    \includegraphics[width=\linewidth]{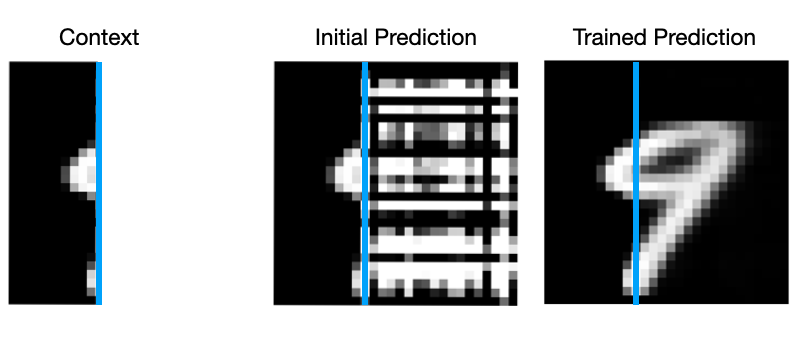}
    \caption{\textbf{Image completion as a multivariate forecasting problem.} \method\hspace{1mm} learns to complete images given the leftmost columns.}
    \label{fig:mnist_example}
\end{figure}

\method's architecture and embedding scheme prompt some interesting parallels with work on Transformers in computer vision. Both domains involve two dimensional data (rows/columns of pixels in vision and space/time in forecasting). In fact, another way to look at the standard multivariate forecasting problem (Sec. \ref{sec:temporal_related}) is as the completion of the rightmost columns of a grayscale image given the leftmost columns. We let $x$ be the scalar index of a column and $y$ be the vector of pixel values for that column. The number of variables corresponds to the number of rows in the image. An example of using \texttt{Spacetimeformer} to complete MNIST images given the first $10$ columns of pixels is shown in Figure \ref{fig:mnist_example}. Our model solves this problem with the same approach used in all the other forecasting results presented in this paper. The image-completion perspective can be an intuitive way to identify the problem with standard ``temporal" attention in TSF Transformers. If faced with the image completion task in Figure \ref{fig:mnist_example}, we would be unlikely to try and model all of the rows of the image together and perform attention over columns - but that is exactly what models like \texttt{Informer} end up doing. Pixel shapes can have complex local structure and each region should have the freedom to attend to distinct parts of the image that are most relevant. We run similar spatiotemporal (\texttt{Spacetimeformer}) and temporal (\texttt{Temporal}) architectures on MNIST and see an $8\%$ reduction in prediction error. However, MNIST is a simple dataset and we would expect a much larger gap in performance on higher-resolution images.

The key difference between image and time series data is that both dimensions of vision data have meaningful order while the order of our spatial axis must be assumed to be arbitrary. This can limit our ability to apply common vision techniques to reduce sequence length. For example, the original Vision Transformer \cite{dosovitskiy2020image} established a convention of ``patching" the image into a grid of small ($16 \times 16$) pixel squares. The input to the Transformer then becomes a feed-forward layer's projection of regions of pixels rather than every pixel individually - dramatically reducing the input length of the self-attention mechanism. Square patches of multivariate time series data would arbitrarily group multiple variables together based on the order their columns appeared in the dataset. However, we do have the option to patch along rectangular ($1 \times k$) slices of time. The initial convolution approach (Sec. \ref{sec:method:scaling}) used on large datasets can be seen as a kind of overlapping patching to reduce sequence length. 

Another common theme in Vision Transformer work is the combination of attention and convolutions. Convolutions provide an architectural bias towards groups of adjacent pixels while attention allows for global connections even in shallow layers where convolutional receptive fields are small \cite{li2021localvit}. \texttt{Spacetimeformer} ``intermediate convolutions" (Sec. \ref{sec:method:scaling}) rearrange the flattened spatiotemporal attention sequence to a Conv1D input format to get a similar effect. Some Vision Transformers create a bias towards nearby pixels with efficient attention mechanisms that resemble convolutions over local regions but form sparse hierarchies across the full image \cite{dong2021cswin, chu2021twins, wang2021pyramid}. \texttt{Spacetimeformer}'s local attention layers can be interpreted as a version of this approach. However, our model also contains true global layers that would be the equivalent of attention between each pixel and every other pixel in an image - something that is not usually attempted in vision architectures. An interesting technique related to the local vs. global approach and convolutional networks is Shifted Window Attention \cite{swin_transformer}. Attention is performed only amongst the pixels or patches of a ``window" or neighborhood of pixels, but the window boundaries are redrawn in each layer so that information can spread between distant windows as the network gets deeper. This is directly analogous to the expanding receptive field that results from down-sampled convolutional architectures. \texttt{Spacetimeformer} implements shifted windowed attention in one dimension where neighborhoods of data are defined by slices in time. This mechanism is not used in the primary experimental results because fast attention provides sufficient memory savings. In general, the overlap between vision and time series techniques appears to be a promising direction for the future scalability of spatiotemporal forecasting. Our public codebase provides fast setup for time series models to train on the MNIST task in Fig. \ref{fig:mnist_example}. We also include a more difficult CIFAR-10 task where the images' rows and columns have been flattened into a sequence with three variables corresponding to the red, green, and blue color channels.

\section{Additional Results and Details}

\subsection{Real-World Dataset Details}
\label{app:dataset_details}

Details of the real-world datasets used in our experimental results are listed below. Descriptions of toy datasets are deferred to the appendix sections where their results are discussed. We try to follow the train/val/test splits of prior work where possible to ensure a fair comparison. Splits are often determined by dividing the series by a specific point in time, so that the earliest data is the train set and the most recent data is the test set. This evaluation scheme helps measure the future predictive power of the model and is especially important in non-stationary datasets \cite{cerqueira2020evaluating}. Datasets are either released directly with our source code or made available for download in the proper format.

\begin{table}[h!]
\centering
\begin{tabular}{@{}lccc@{}}
\toprule
                 & \textbf{\begin{tabular}[c]{@{}c@{}}Variables \\ ($N$)\end{tabular}} & \textbf{\begin{tabular}[c]{@{}c@{}}Length \\ ($L$)\end{tabular}} & \textbf{\begin{tabular}[c]{@{}c@{}}Size \\ (Timesteps)\end{tabular}} \\ \midrule
NY-TX Weather    & 6                                                                   & 800                                                              & 569,443                                                              \\
AL Solar         & 137                                                                 & 192                                                              & 52,560                                                               \\
Metr-LA Traffic  & 207                                                                 & 24                                                               & 34,272                                                               \\
Pems-Bay Traffic & 325                                                                 & 24                                                               & 52,116 \\ 
ETTm1 & 7 & 1344 & 69,680 \\
HZMetro & 160 & 8 & 1650  \\
Weather & 21 & 1440 & 52,696
\\ \bottomrule
\end{tabular}
\caption{\textbf{Real-World Dataset Summary.} Sequence length ($L$) is reported as the largest combined length of the context and target windows used in results.}
\label{tbl:dataset_summary}
\end{table}

\begin{itemize}

    \item \textbf{NY-TX Weather (\textit{new})}. We obtain hourly temperature readings from the ASOS weather network. We use three stations located in central Texas and three more located hundreds of miles away in eastern New York. The data covers the years $1949-2021$, making this a very large dataset by TSF standards. Many of the longest active stations are airport weather stations. We use the airport codes ACT (Waco, TX), ABI (Abilene, TX), AMA (Amarillo, TX), ALB (Albany, New York) as well as the two largest airports in the New York City metro, LGA and JFK. The NY-TX benchmark was created for this paper, and has two key advantages when evaluating timeseries Transformers. First, it has far more timesteps than popular datasets; Transformers are data-hungry, but we can more safely ignore overfitting issues here.  Second, this dataset is almost perfectly stationary; popular datasets are not, and this creates an evaluation problem that is not related to the modeling power of the architecture. This issue has been widely overlooked and allows basic linear models that are more robust to non-stationary to outperform recent methods that are more powerful on paper. For more on this see \cite{aretransformerseffective} and Appendix \ref{app:ettm1_appendix}. 
    \item \textbf{Metr-LA Traffic \cite{li2018diffusion}}. A popular benchmark dataset in the GNN literature consisting of traffic measurements from Los Angeles highway sensors at $5$-minute intervals over $4$ months in $2012$. Both the context and target sequence lengths are set to $12$. We use the same train/test splits as \cite{li2018diffusion}.
    \item \textbf{AL Solar \cite{lai2018modeling}}. A popular benchmark dataset in the time series literature consisting of solar power production measurements across the state of Alabama in $2006$. We use a context length of $168$ and a target length of $24$.
    \item \textbf{Pems-Bay Traffic \cite{li2018diffusion}}. Similar to Metr-LA but covering the Bay Area over $6$ months in $2017$. The context and target lengths are $12$ and we use the standard train/test split.
    \item \textbf{HZMetro \cite{liu2020physical}}. A dataset of the number of riders arriving and departing from $80$ metro stations in Hangzhou, China during the month of January $2019$. The $(L \times N, 1)$ input format of our model requires the arrivals and departures to be separated into their own variable nodes, leading to a total of $160$ variables. The context and target windows are set to a length of $4$ as in previous work.
    \item \textbf{ETT \cite{zhou2021informer}}. An electricity transformer dataset covering $2016-2018$. We use the ETTm1 variant which is logged at one minute intervals to provide as much data as possible. We evaluate on the set of target sequence lengths $\{24, 48, 96, 288, 672\}$ established by \texttt{Informer}.
    \item \textbf{Weather \cite{zhou2021informer}}. A German dataset of $21$ weather-related variables recorded every $10$ minutes in $2020$. We use the same set of target sequence lengths and context window selection approach as in ETTm1.
\end{itemize}

\subsection{Code and Implementation Details}
\label{app:implementation}
The code for our work is open-sourced and available on GitHub at \href{https://github.com/QData/spacetimeformer}{QData/spacetimeformer}. All models are implemented in PyTorch \cite{NEURIPS2019_9015} and the training process is conducted with PyTorch Lightning \cite{falcon2019pytorch}. Our \texttt{LSTNet} and \texttt{MTGNN} implementations are based on public code \cite{rozemberczki2021pytorch} and verified by replicating experiments from the original papers. Generic models like \texttt{LSTM} and \texttt{LinearAR} are implemented from scratch and we made an effort to ensure the results are competitive. The code for the \texttt{Spacetimeformer} model was originally based on the \texttt{Informer} open-source release.

\textbf{Data Preprocessing}.  As mentioned in the previous section, train/val/test splits are based on existing work or determined by a temporal splits where the most recent data forms the test set. Data is organized into time sequences ($x$) and variable values ($y$). We follow established variable normalization schemes of prior work to ensure a fair comparison, and default to z-score normalization in other cases. Most real-world datasets use $x$ values that correspond to date/time information. We represent calendar dates by splitting the information into separate year, month, day, hour, minute, and second values and then re-scaling each to be $\in [0, 1]$. This works out so that only the year value is unbounded, but we divide by the latest year present in the training set. We discard time variables that are not robust or prone to overfitting. For example, a dataset that only spans two months would drop the month and year values.

\textbf{Embedding}. The time variables $x$ are passed through a \texttt{Time2Vec} layer \cite{kazemi2019time2vec}. If $x$ is a time representation with three elements \{hour, minute, second\}, the \texttt{Time2Vec} output would be shape $(3,  k)$ where $k$ is the time embedding dimension. The first of the $k$ elements has no activation function while the remaining $k-1$ use a sine function with trainable parameters to represent periodic patterns. After flattening $(3, k) \rightarrow (3 \times k,)$, the time embedding is concatenated with its corresponding $y$ value. When using our spatiotemporal embedding, the y value will be a single scalar and time values are duplicated to account for the flattened sequence. Note that we use terminology like ``flatten" because most datasets in practice are structured such that there are $N$ variables sampled at the same moment in time, and the conversion to the spatiotemporal format looks like we have laid out the rows of a dataframe end-to-end. However, embedding the value of each variable at every timestep as its own separate token gives us a lot of flexibility to use alternate dataset formats where variables may be sampled at different intervals. We do not take advantage of this in our experimental results because it is not relevant to common benchmark datasets and is not applicable to all the baselines we consider.

The combined value and time are projected to the Transformer model dimension ($d$) with a single feed-forward layer. We experimented with two approaches to the position embedding. The first re-purposes \texttt{Time2Vec} to learn a periodic $d$-dimensional embeddings - essentially learning the frequency hyperparameters of the original fixed position embedding \cite{vaswani2017attention} automatically. We also experimented with fully learnable lookup-table-style position embeddings commonly used for word embeddings in natural language processing. Both are provided in the code and appeared to lead to similar performance. However, we decided that the fully learnable option was the safer choice to make sure the position embedding had the freedom to differentiate itself from the other components that make up our embedding scheme. Position indices are repeated as necessary to account for the flattened $y$ values. Variable embeddings are implemented similarly with indices assigned arbitrarily from $0, \dots, N$. The repeating pattern of tokens' variable and position indices is best explained by Figure \ref{fig:embed_seq}.

``Given" embeddings are a third lookup-table layer with two entries that indicate whether the $y$ value for a token contains missing values. They are only used in the encoder's embedding layer because all of the values are missing/empty in the decoder sequence. The value+time, variable, position, and given embeddings sum to create the final $d$-dimensional embedding.

\textbf{Architecture and Training Loop.} There is significant empirical work investigating technical improvements to the Transformer architecture and training routine \cite{pmlr-v119-huang20f, liu2020understanding, wu2021unidrop, zhou2020scheduled, araabi2020optimizing, fan2019reducing}. We incorporate some of these techniques to increase performance and hyperparameter robustness while retaining simplicity. A Pre-Norm architecture \cite{xiong2020layer} is used to forego the standard learning rate warmup phase. We also find that replacing \texttt{LayerNorm} with \texttt{BatchNorm} is advantageous in the time series domain. \cite{shen2020powernorm} argue that \texttt{BatchNorm} is more popular in computer vision applications because reduced training variance improves performance over the \texttt{LayerNorm}s that are the default in NLP. Our experiments add empirical evidence that this may also be the case in time series problems. We also experiment with \texttt{PowerNorm} \cite{shen2020powernorm} and \texttt{ScaleNorm} \cite{nguyen2019transformers} layers with mixed results. All four variants are included as a configurable hyperparameter in our open-source release.

The encoder and decoder have separate embedding layers to enable the context sequence to contain a different set of input variables than those we are forecasting in the target sequence. Dropout can be applied to: 1) query, key, and value layer parameters. 2) The output of embedding layers. 3) The attention matrix (when applicable). 4) The feed-forward layers at the end of each encoder/decoder layer. Local attention layers are implemented with \texttt{rearrange}-style operations \cite{rogozhnikov2022einops} and can be used with any efficient attention variant. The choice of attention implementation is flexible across the architecture, allowing progress in the sub-field of long-range attention to improve future scalability. We default to the ReLU version of \texttt{Performer} \cite{choromanski2021rethinking} due to its memory savings and compatibility with both self and cross attention. However, we also experimented with \texttt{Nystromformer} \cite{xiong2021nystromformer} and \texttt{ProbSparse} attention \cite{zhou2021informer}.

\textbf{Metrics.} For completeness, the evaluation metrics used in our results tables are listed below.  $y_t^n$ and $\hat{y}_t^n$ correspond to the true and predicted values of the $n$th variable at timestep $t$, respectively. $\Bar{y}$ is shorthand for the mean value of $y$. 

\begin{align}
    \text{MSE} &:= \frac{1}{hN}\sum_{n=1}^{N}\sum_{t=T}^{T+h}(y_{t}^n - \hat{y}_{t}^n)^2 \\
    \text{MAE} &:= \frac{1}{hN}\sum_{n=1}^{N}\sum_{t=T}^{T+h}(y_{t} - \hat{y}_{t}) \\
     \text{RMSE} &:= \sqrt{\frac{1}{hN}\sum_{n=1}^{N}\sum_{t=T}^{T+h}(y_{t} - \hat{y}_{t})^2} \\
     \text{RRSE} &:= \sqrt{\frac{\sum_{n=1}^{N}\sum_{t=T}^{T+h}(y_t - \hat{y}_t)^2}{\sum_{n=1}^{N}\sum_{t=T}^{T+h}(y_t - \Bar{y})^2}} \\
     \text{MAPE} &:= \frac{1}{hN}\sum_{n=1}^{N}\sum_{t=T}^{T+h}\left|\frac{(y_t - \hat{y}_t)}{y_t}\right|
\end{align}

The main detail to note here is that we are reporting the average over the length of the forecast sequence of $h$ timesteps. However, it is somewhat common (especially in the ST-GNN datasets) to report metrics for multiple timesteps independently to see how accuracy changes as we get further from the known context. We take the mean of all reported timesteps to recover the metrics of related work for Tables \ref{tbl:traffic_results} and \ref{tbl:metro_results}. All metrics discard missing values.

As mentioned in Sec \ref{sec:arch}, the output of our model is a sequence of predictions that can be restored to the original ($L, N$) input format. This lets us choose from several different loss functions. Most datasets use either Mean Squared Error (MSE) or Mean Absolute Error (MAE). We occasionally compare Root Relative Squared Error (RRSE) and Mean Absolute Percentage Error (MAPE) - though they are never used as an objective function. We train all models with early stopping and learning rate reductions based on validation performance.

\textbf{Hyperparameters and Compute.}
Due to the large number of experiments and baselines used in our results, we choose to defer hyperparamter information to the source code by providing the necessary training commands. This lets us include the settings of minor details that have not been explained in writing. \texttt{Spacetimeformer}'s long-sequence attention architecture is mainly constrained by GPU memory. Most results in this paper were gathered on $1-4$ $12$GB cards, although larger A$100$ $80$GB GPUs were used in some later results (e.g., Pems-Bay). We hope to provide training commands that serve as a competitive alternative for resource-constrained settings. Training is relatively fast due to small dataset sizes, with the longest results being Metr-LA and Pems-Bay at roughly $5$ hours.

\subsection{Toy Experiments}
\label{app:toy_exps}

\begin{figure}[h!]
    \centering
    \includegraphics[width=1.1\linewidth]{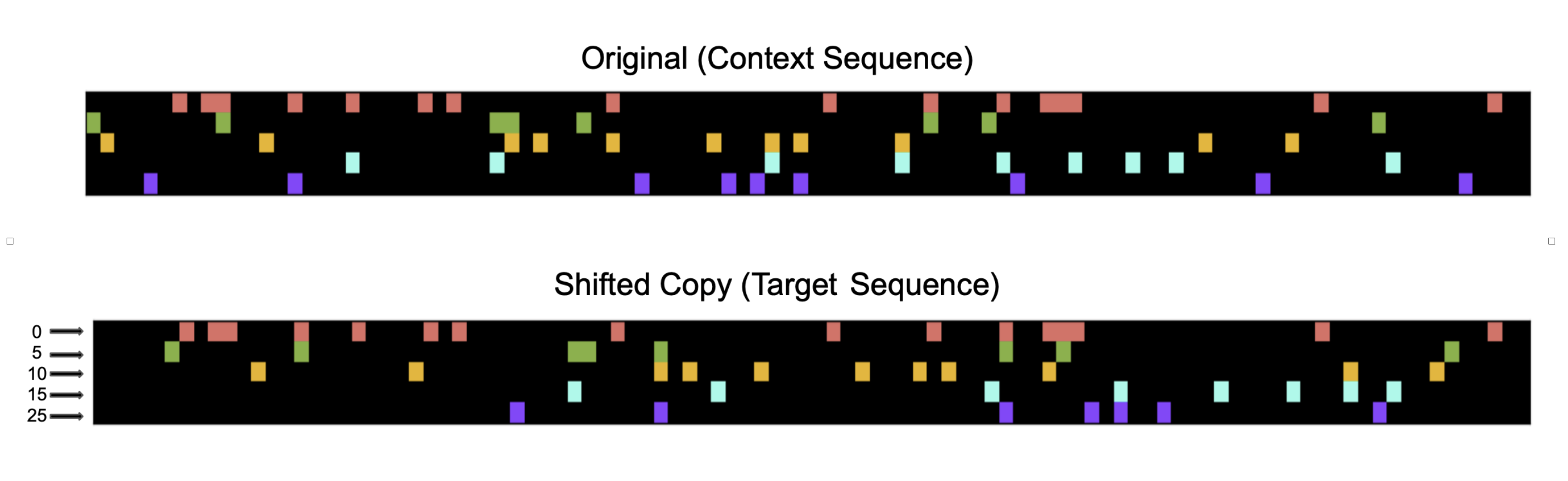}
    \caption{\textbf{Shifted Copy Task.} Copy a binary sequence with each row shifted by increasing amounts top to bottom (colorized for visualization).}
    \label{fig:shifted_copy}
\end{figure}

\textbf{Shifted Copy Task.} Copying binary input sequences is a common test of long-sequence or memory-based models \cite{graves2014neural}. We generate binary masks with shape $(L, N)$, where elements are set to $1$ with some probability $p$. Each of the $N$ variables is associated with a ``shift" value. The target sequence is a duplicate of the context sequence with each variable zero-padded and translated forward in time by its shift value. An example with $L=100$, $N=5$, and shifts of $\{0, 5, 10, 15, 25\}$ is shown in Figure \ref{fig:shifted_copy}. The rows have been color-coded to make the shift easier to identify.

\begin{figure}[h!]
    \centering
    \includegraphics[width=1.05\linewidth]{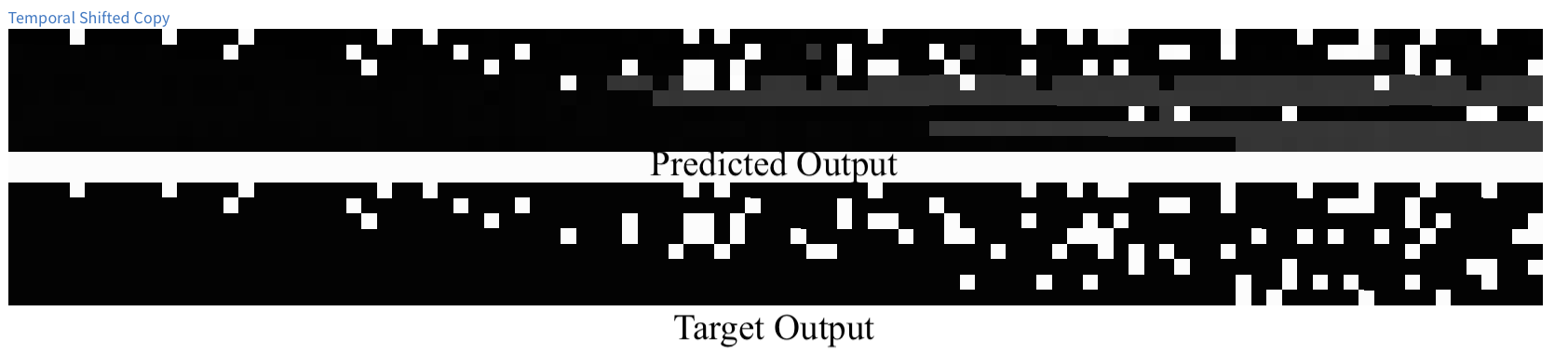}
    \caption{\textbf{\texttt{Temporal} attention discrete copy example.} Top image shows predicted output sequence; bottom shows ground-truth output sequence. Standard attention forces each column to share attention graphs, leading to a ``blurry" copy output on variables with high shift.}
    \label{fig:shifted_copy_temporal_patterns}
\end{figure}

First we train a standard (``\texttt{Temporal}") Transformer on an eight variable version of the shifted copy task. An example of the correct output (bottom) and predicted sequence (top) is shown in Figure \ref{fig:shifted_copy_temporal_patterns}. While the first few rows appear to be accurately copied, the lower variable outputs are blurry or missing entirely.

\begin{figure*}
    \centering
    \includegraphics[width=1\textwidth]{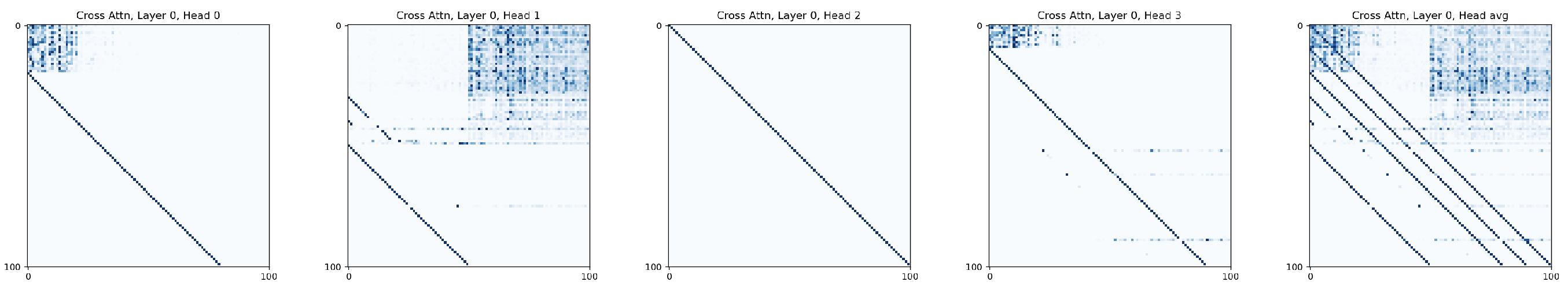}
    \caption{\textbf{\texttt{Temporal} attention fails with too few heads.} Because each variable (row) requires an independent attention graph, a strong optimizer will attempt to put one relationship on each attention head. However, we quickly run out of heads and are left with an inaccurate copy. In a more complex problem, Temporal attention could require as many as $N^2$ heads to accurately model every variable relationship.}
    \label{fig:shifted_copy_temporal}
\end{figure*}

To try and understand where the model is going wrong, we visualize the cross attention patterns of several heads in the first decoder layer. The results are shown in Figure \ref{fig:shifted_copy_temporal}. To perform this task correctly, the tokens of each element in the target sequence need to learn to attend to a specific position in the context sequence. The problem is that each variable needs to attend to different positions due to their temporal shift. The standard Transformer groups all of the variables into the same token, making it very difficult to pick timesteps to attend to that are useful for all of them. Instead it correctly learns the shift of a single variable per attention head and eventually runs out of heads, leading to the inaccurate results in the bottom half of the output. Note that in this problem the correct attention pattern would form a line where the distance from the diagonal corresponds to a variable's shift value. 

\begin{figure}
    \centering
    \includegraphics[width=\linewidth]{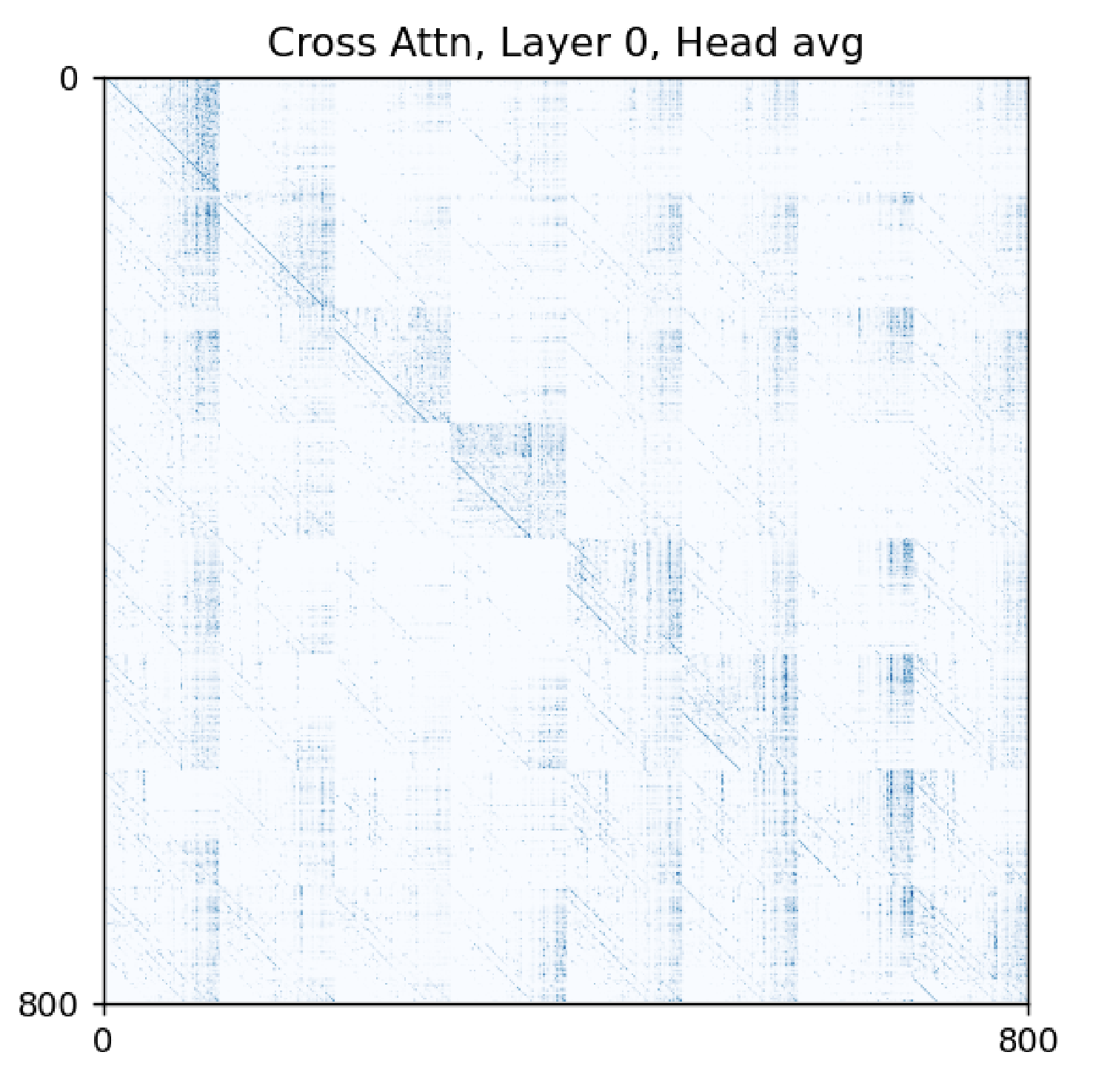}
    \caption{\textbf{Spatiotemporal attention represents $N^2$ relationships per head.} Thanks to its spatiotemporal sequence embedding, \method \hspace{1mm} can represent the underlying variable relationship in a single accurate head. This capability spares room for optimization errors and multiple relationships between variables in more complex problems.}
    \label{fig:shifted_copy_spatiotemporal_patterns}
\end{figure}

\begin{figure}
    \centering
    \includegraphics[width=\linewidth]{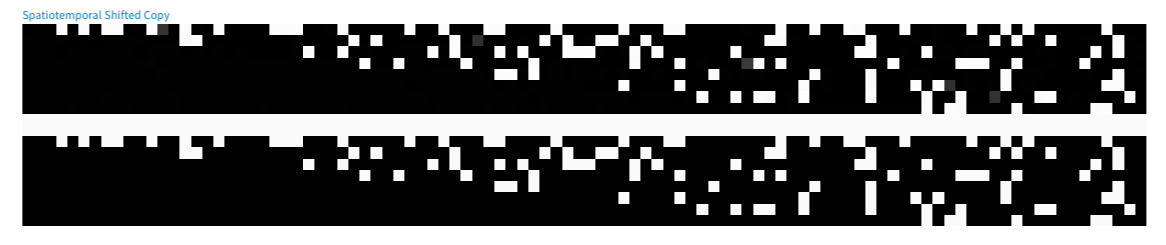}
    \caption{\textbf{\method \hspace{1mm} outputs an accurate copy (top) of the input sequence (bottom) despite the row shift.}}
    \label{fig:shifted_copy_spatiotemporal}
\end{figure}

Next we run the same Transformer with spatiotemporal attention and see a \textit{roughly $17\times$ reduction in MAE}, with outputs that are near-perfect reproductions of the input (Figure \ref{fig:shifted_copy_spatiotemporal_patterns}). By flattening the context and target tokens into a spatiotemporal graph, the variables of each timestep are given independent attention mechanisms. Each attention head is capable of learning $N^2$ relationships between variables. Of course, this toy problem only has $N$ important spatial relationships (each variable only needs to attend to itself). Figure \ref{fig:shifted_copy_spatiotemporal} shows the spatiotemporal pattern with attention shifting along the diagonal. \\

\textbf{Dependent Sine Waves.} Next we recreate a version of the toy dataset used to emphasize the necessity of spatial modeling in \cite{shih2019temporal}. We generate $D$ sequences where sequence $i$ at timestep $t$ is defined by:
    
    \begin{equation}
    Y_{t}^{i} = sin(\frac{2\pi i t}{64}) + \frac{1}{D+1}\sum_{j=1, j\neq i}^{D}sin(\frac{2\pi j t}{64})
    \end{equation}
    
We map $2,000$ timesteps to a sequence of daily calendar dates beginning on Jan 1, $2000$. We set $D=20$ and use a context length of $128$ and a target length of $32$. The final quarter of the time series is held out as a test set.

Several ablations of our method are considered. \texttt{Temporal} modifies the spatiotemporal embedding as discussed at the end of Sec \ref{sec:arch}. \texttt{ST Local} skips the global attention layers but includes spatial information in the embedding. The ``Deeper" variants attempt to compensate for the additional parameters of the local+global attention architecture of our full method. All models use a small Transformer model and optimize for MSE. The results are shown in Table \ref{tbl:toy_results}. The \texttt{Temporal} embedding is forced to compromise its attention over timesteps in a way that reduces predictive power over variables with such distinct frequencies. Standard (``Full") attention fits in memory with the Temporal embedding but is well approximated by \texttt{Performer}. Our method learns an uncompromising spatiotemporal relationship among all tokens to generate the most accurate predictions by all three metrics.

\begin{table}[h!]
\resizebox{\columnwidth}{!}{
\begin{tabular}{@{}ccccccc@{}}
\toprule
             & \textbf{Temporal} & \begin{tabular}[c]{@{}l@{}}\textbf{Temporal} \\ \textbf{(Deeper)}\end{tabular} & \begin{tabular}[c]{@{}l@{}}\textbf{Temporal}\\ \textbf{(Deeper} \&\\ \textbf{Full Attn)}\end{tabular} & \textbf{ST Local} & \begin{tabular}[c]{@{}l@{}}\textbf{ST Local}\\  \textbf{(Deeper)}\end{tabular} & \begin{tabular}[c]{@{}l@{}} \method \end{tabular} \\ \midrule
MSE & 0.006    & 0.010                                                        & 0.005                                                                 & 0.021    & 0.014                                                        & \underline{0.003}                                                \\
MAE          & 0.056    & 0.070                                                        & 0.056                                                                  & 0.104    & 0.090                                                        & \underline{0.042}                                                \\
RRSE         & 0.093    & 0.129                                                        & 0.094                                                                  & 0.180    & 0.153                                                        & \underline{0.070}                                                \\ \bottomrule
\end{tabular}
}
\caption{\textbf{Toy Dataset Results}}
\label{tbl:toy_results}
\end{table}

\subsection{Distributional Shift and Time Series Transformers}
\label{app:ettm1_appendix}

While collecting baseline results on ETT and other datasets used in recent Transformer TSF work we noticed significant gaps in performance between train and test set prediction error. However, the apparent overfitting effect showed few signs of improvement when given additional data and smaller, highly-regularized architectures. Recently, \cite{aretransformerseffective} released an investigation into the failures of Transformers in long-term TSF. They show that a simple model, ``\texttt{DLinear}," can outperform Transformers in a variety of common benchmarks. Their \texttt{DLinear} model is a slightly more advanced version of our \texttt{LinearAR} baseline, which predicts each timestep as a linear combination of the previous timesteps. We verify a similar result with the tiny \texttt{LinearAR} model in Table \ref{tab:ettm1more} - nearly matching the performance of \texttt{ETSFormer} on the ETTm1 dataset.  

Comparisons between the NY-TX Weather dataset - where \texttt{LinearAR} performs worse than our Transformer baselines - and ETTm1 show the latter dataset has a much larger gap between the magnitude of the train and test sequences. This type of distributional shift is common in non-stationary time series and is caused by the use of the most recent data as a test set; long-term trends that begin in the train set can continue into the test set and alter the distribution of our inputs. We hypothesize that distribution shift is a key part of the Transformer's test set inaccuracy and that linear models are less effected because their outputs are a more direct combination of the magnitude of their inputs. Reversible Instance Normalization (\texttt{RevIN}) \cite{kim2021reversible} normalizes a model's input based on the statistics of each context sequence so that its parameters are less sensitive to changes in scale. The predicted output can then be inverse normalized to the original range. We experiment with \texttt{RevIN} as a way to combat distribution shift on both ETTm1 and the popular Weather dataset. The results are displayed in Tables \ref{tab:ettm1more} and \ref{tbl:weather}. Input normalization improves the performance of all models and even makes LSTM competitive with advanced Transformers. This may suggest that the improved performance of recent methods may not be the result of more complex architectures but from improved resistance to distributional shift. We investigate this further by implementing seasonal decomposition - another common detail of Transformers in time series. Seasonal decomposition separates a series into short and long-term trends to be processed independently, and is also included in the \texttt{DLinear} model \cite{aretransformerseffective}. We evaluate its impact on the Weather dataset in Table \ref{tbl:weather}. Performance is comparable to instance normalization, potentially because decomposition has the effect of standardizing inputs by transforming them into a difference from a moving average. The \texttt{Spacetimeformer} results in Table \ref{tbl:ettm1} use both seasonal decomposition and reversible instance normalization for further improved results.

However, input normalization is not always enough, especially when patterns are in fact scale-dependent. For a brief example consider a continuous version of the copy task in the previous section using sine curves instead of binary masks. We randomly generate $N$ different context and target sequences according to:

\begin{align*}
    y_{(t)} &= \textcolor{blue}{a} \text{ sin}(\textcolor{blue}{b}t + \textcolor{blue}{c}) + \textcolor{red}{d} + \epsilon(t), \epsilon(t) \sim \mathcal{N}(0, .1) \\
\end{align*}
where $a, b, c$ and $d$ are parameters drawn from distributions that can vary between the train and test sets. Our experiments show that \texttt{RevIN}-style input normalization can reduce the generalization gap for out-of-distribution $d$ to nearly zero for LSTM, temporal and spatiotemporal attention models. However, if we shift the curve between the context and target such that:

\begin{align*}
 y_{\text{context}}(t) &= \textcolor{blue}{a} \text{ sin}(\textcolor{blue}{b}t + \textcolor{blue}{c}) + \textcolor{red}{d} + \epsilon(t), \epsilon(t) \sim \mathcal{N}(0, .1) \\
 y_{\text{target}}(t) &= y_{context}(t) + \underline{(1 + |\textcolor{red}{d}|)^2}
\end{align*}

\begin{figure}[h!]
    \centering
    \includegraphics[width=\linewidth]{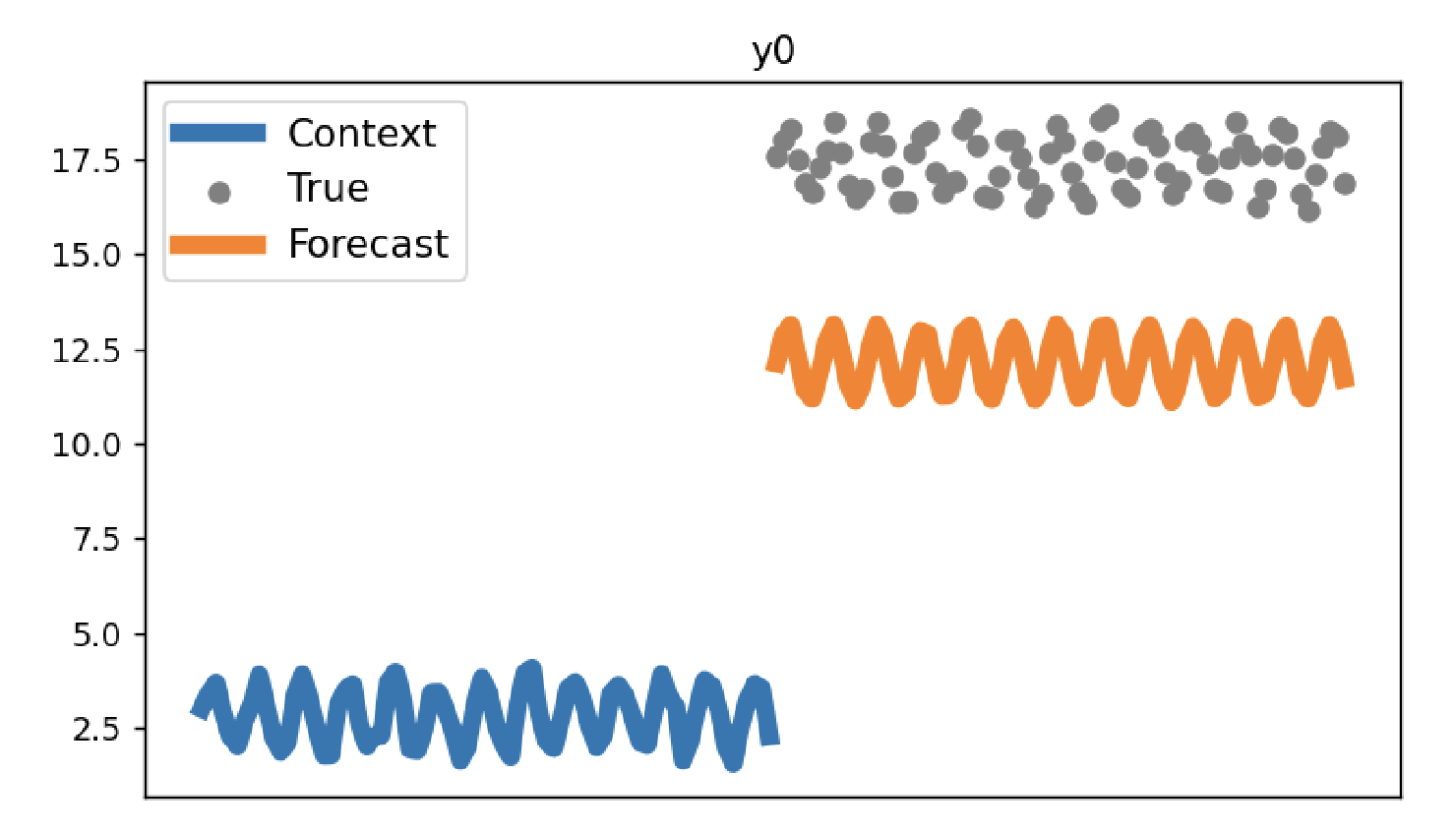}
    \caption{\textbf{Modeling Without Magnitude.} Input normalization removes the context information that is needed to make scale-dependent predictions.}
    \label{fig:case2}
\end{figure}

RevIN is unable to generalize. This is because when we pass our model a normalized input we are discarding the magnitude ($d$) information necessary to make accurate predictions. An example of a \texttt{Spacetimeformer+RevIN} prediction is plotted in Figure \ref{fig:case2}. The behavior of many real-world time series also changes with the overall magnitude of the variable and more work is needed to make large Transformer TSF models robust to small, non-stationary datasets.

\begin{table}[]
\resizebox{\linewidth}{!}{
\begin{tabular}{cccccc}
\hline
                                   & \multicolumn{5}{c}{Prediction Length}                                                                                       \\
                                   & 24                          & 48                          & 96                          & 288                         & 672                         \\ \hline
LSTM                      & 0.63                        & 0.94                        & 0.91                        & 1.12                        & 1.56                        \\
{ Repeat Last} & { 0.54} & { 0.76} & { 0.78} & { 0.83} & { 0.87} \\
Informer                   & 0.37                        & 0.50                        & 0.61                        & 0.79                        & 0.93                        \\ \hline
Pyraformer                         &                             &                             & 0.49                        & 0.66                        & 0.71                        \\
YFormer                           & 0.36                        & 0.46                        & 0.57                        & 0.59                        & 0.66                        \\
Preformer                          & 0.40                        & 0.43                        & 0.45                        & 0.49                        & 0.54                        \\
Autoformer                         & 0.40                        & 0.45                        & 0.46                        & 0.53                        & 0.54                        \\
ETSFormer                & 0.34                        & 0.38                        & 0.39                        & {\ul 0.42}                  & {\ul 0.45}                  \\ \hline
LSTM + RevIN              & 0.37                        & 0.44                        & 0.47                        & 0.53                        & 0.55                        \\
Temporal + RevIN          & {\ul 0.32}                  & 0.40                        & 0.44                        & 0.50                        & 0.55                        \\
Spatiotemporal + RevIN    & 0.34                        & 0.40                        & 0.45                        & 0.50                        & 0.57                        \\ \hline
LinearAR                 & 0.33                        & {\ul 0.37}                  & {\ul 0.39}                  & 0.44                        & 0.48                        \\ \hline
\end{tabular}
}
\caption{Normalized Mean Absolute Error (MAE) on ETTm1 test set. ``Repeat Last" is a simple heuristic that predicts the most recent value in the context sequence for the duration of the forecast.}
\label{tab:ettm1more}
\end{table}

\begin{table}[]
\begin{tabular}{@{}lcccc@{}} \toprule
 & \multicolumn{4}{c}{Prediction Length} \\ \midrule
 & 96 & 192 & 336 & 720 \\ \midrule
Informer & 0.38 & 0.54 & 0.52 & 0.74 \\
LSTM (from Autoformer) & 0.41 & 0.44 & 0.45 & 0.52 \\
Autoformer & 0.34 & 0.37 & 0.40 & 0.43 \\ \midrule
Linear Shared & 0.24 & 0.29 & 0.32 & 0.36 \\
Linear Ind Decomp & 0.22 & 0.27 & 0.31 & 0.36 \\ \midrule
LSTM (Ours) & 0.30 & 0.34 & 0.41 & 0.53 \\
LSTM RevIN & 0.23 & 0.26 & 0.32 & 0.36 \\
LSTM Decomp & 0.23 & 0.27 & 0.30 & 0.36 \\ \midrule
Transformer (Ours) & 0.24 & 0.33 & 0.40 & 0.42 \\
Transformer RevIN & 0.23 & 0.27 & 0.35 & 0.37 \\
Transformer Decomp & 0.24 & 0.27 & 0.31 & 0.37 \\ \bottomrule
\end{tabular}
\caption{Normalized Mean Absolute Error (MAE) on the Weather test set. ``Linear Ind Decomp" adds independent parameters for each variable to a linear model with seasonal decomposition.}
\label{tbl:weather}
\end{table}

\subsection{Ablations}
\label{app:ablations}
Ablation results are listed in Table \ref{tbl:ablation_results}. During our development process, we became interested in the amount of spatial information that makes it through the encoder and decoder layers. We created a way to measure this by adding a softmax classification layer to the encoder and decoder output sequences. This layer interprets each token and outputs a prediction for the time series variable it originated from. Importantly, we detach this classification loss from the forecasting loss our model is optimizing; only the classification layer is trained on this objective. Classification accuracy begins near zero but spikes upwards of $99\%$ in all problems within a few hundred gradient updates due to distinct (randomly initialized) variable embeddings that are well-preserved by the residual Pre-Norm Transformer architecture. In TSF tasks like NY-TX weather, this accuracy is maintained throughout training. However, in more spatial tasks like Metr-LA, accuracy begins to decline over time. Because this decline corresponds to increased forecasting accuracy, we interpret this as a positive indication that related nodes are grouped with similar variable embeddings that become difficult for a single-layer classification model to distinguish.

The relative importance of space and time embedding information changes based on whether the problem is more temporal like NY-TX Weather or spatial like traffic forecasting. While global spatiotemporal attention is key to state-of-the-art performance on Metr-LA, it is interesting to see that local attention on its own is a competitive baseline. We investigate this further by removing both global attention and the variable embeddings that differentiate traffic nodes, which leads to the most inaccurate predictions. This implies that the local attention graph is adapting to each node based on the static spatial information learned by variable embeddings. 

\begin{table}[th]
\resizebox{\linewidth}{!}{
\begin{tabular}{@{}lcc@{}}
\toprule
                    & MAE         & \begin{tabular}[c]{@{}c@{}}Classification Acc.\\ (\%)\end{tabular} \\ \midrule
\textbf{NY-TX Weather ($L = 200, N = 6$)}       & \multicolumn{1}{l}{} & \multicolumn{1}{l}{}                                               \\ \midrule
Full Spatiotemporal               & \underline{2.57}                 & \underline{99}                                                                \\
No Local Attention            & 2.62                 &  \underline{99}                                                                  \\
No Variable Embedding           & 2.66                 & 45                                                                 \\
Temporal Embedding/Attention            & 2.67                 & -                                                                  \\
No Value Embedding            & 3.83                 & \underline{99}                                                                \\
No Time Embedding             & 4.01                 & \underline{99}                                                                \\ \midrule
\textbf{Metr-LA Traffic ($L = 24, N = 207$)}    &                      &                                                                    \\ \midrule
Full Spatiotemporal                & \underline{2.83}                 & \underline{58}                                                                 \\
No Global Attention          & 3.00                 & 46                                                                 \\
No Time Embedding            & 3.11                 & 50                                                                 \\
Temporal Embedding/Attention           & 3.14                 & -                                                                  \\
No Local Attention           & 3.27                 & 54                                                                 \\
No Variable Embedding           & 3.29                 & 2                                                                  \\
No Global Attention, No Variable Emb. & 3.48                 & 1                                                                  \\ \bottomrule
\end{tabular}
}
\caption{\textbf{Ablation Results.}}
\label{tbl:ablation_results}
\end{table}

\subsection{Future Directions in Multi-series Generalization}
\label{app:multiseries}
Some time series prediction problems - particularly competitions like the M series \cite{MAKRIDAKIS202054} and Kaggle - are based on a large collection of semi-independent series that do not occur on the same time interval and therefore cannot be cast in the standard multivariate format. For example, we might have the sales data of a collection of thousands of different products sold in different stores several years apart. ML-based solutions use a univariate context/target window that is trained jointly across batches from every series. Implicitly, these models need to infer the behavior of the current series they are observing from this limited context. Long-sequence Transformers create an opportunity for genuine ``in-context" learning \cite{kirsch2022general, gpt3} in the time series domain, where the context sequence is \textit{all} existing data for a given varible up until the current moment, and the target sequence is \textit{all} future values we might want to predict. A model trained in this format would have all available information to identify the characteristics of the current series and learn generalizable patterns for more accurate forecasting. The main implementation difference here is the varying number of samples per series, which requires the use of sequence padding and masking and can limit our choice of efficient attention variant. For multi-series problems, we revert to vanilla (quadratic) attention with arbitrary masking, and make use of shifted-window attention (Sec. \ref{sec:method:scaling}) to reduce GPU memory usage. Our open-source code release includes fully implemented training routines for the Monash dataset \cite{godahewa2021monash}, M4 dataset \cite{MAKRIDAKIS202054}, and \href{https://www.kaggle.com/c/web-traffic-time-series-forecasting}{Wikipedia site traffic Kaggle competition}. Preliminary results suggest this method can be competitive with leaderboard scores of ML-based approaches on the M4 and Wikipedia competitions, and we hope that our codebase can be a starting point for further development on this topic. The multivariate attention capabilities of \method\hspace{1px} allow an extension of meta-TSF to datasets of multiple multivariate time series.

\end{document}